%% file: main_arxiv.tex
\documentclass{article}

\usepackage[preprint]{neurips_2019}

\usepackage[utf8]{inputenc} 
\usepackage[T1]{fontenc}    
\usepackage{hyperref}       
\usepackage{url}            
\usepackage{booktabs}       
\usepackage{amsfonts}       
\usepackage{nicefrac}       
\usepackage{microtype}      

\usepackage{subcaption}
\usepackage{graphicx}
\usepackage{booktabs} 
\usepackage{amsmath}
\usepackage{amsfonts}
\usepackage{multirow}
\usepackage[table]{xcolor}
\usepackage{algorithm,algpseudocode}
\usepackage[flushleft]{threeparttable}

\usepackage{makecell}

\usepackage{multicol} 
\usepackage{pbox}
\usepackage{float}

 \title{DEEP-BO for Hyperparameter Optimization of Deep Networks}

\author{%
	Hyunghun Cho \\
	Department of Transdisciplinary Studies, \\ Seoul National University \\
	\texttt{webofthink@snu.ac.kr} \\	
	\And
	Yongjin Kim \\
	College of Liberal Studies,\\ Seoul National University \\
	\texttt{yongjinkim@snu.ac.kr} \\
	\AND
	Eunjung Lee, Daeyoung Choi \\
	Department of Transdisciplinary Studies, \\ Seoul National University \\
	\texttt{\{ej-lee,choid\}@snu.ac.kr} \\
	 \And
	Yongjae Lee \\
	School of Management Engineering, \\ UNIST \\
	\texttt{yongjaelee@unist.ac.kr} \\	
	 \And
	Wonjong Rhee \\
    Department of Transdisciplinary Studies, \\ Seoul National University \\
    \texttt{wrhee@snu.ac.kr} \\
}

\begin{document}
	
	\maketitle

\begin{abstract}

The performance of deep neural networks (DNN) is very sensitive to the particular choice of hyper-parameters. To make it worse, the shape of the learning curve can be significantly affected when a technique like batchnorm is used. As a result, hyperparameter optimization of deep networks can be much more challenging than traditional machine learning models. In this work, we start from well known Bayesian Optimization solutions and provide enhancement strategies specifically designed for hyperparameter optimization of deep networks. 
The resulting algorithm is named as DEEP-BO (Diversified, Early-termination-Enabled, and Parallel Bayesian Optimization). When evaluated over six DNN benchmarks, DEEP-BO easily outperforms or shows comparable performance with some of the well-known solutions including GP-Hedge, Hyperband, BOHB, Median Stopping Rule, and Learning Curve Extrapolation.
The code used is made publicly available at \url{https://github.com/snu-adsl/DEEP-BO}.

\end{abstract}

\input{1_introduction.tex}

\input{2_related_works.tex}
\input{3_deep_bo.tex}
\input{4_experiment.tex}
\input{5_discussion.tex}
\input{6_conclusion.tex}

\bibliographystyle{abbrvnat}
\bibliography{diversified_hpo}

\newpage


\input{supple-a.tex}

\newpage


\input{supple-b.tex}

\newpage


\input{supple-c.tex}

\newpage
\input{supple-d.tex}

\end{document}

%% file: 1_introduction.tex
\section{Introduction}
\label{Introduction}

Hyperparameter optimization(HPO) aims to find the global optima $\mathbf{x}^*$ of an unknown black-box function $f$. That is, $\mathbf{x}^*=\operatorname*{arg\,max}_{\mathbf{x} \in \mathcal{X}} f(\mathbf{x})$, where $\mathcal{X}$ is a hyperparameter space that can be a subset of $\mathbb{R}^d$, contain categorical, or discrete variables. $f(\mathbf{x})$ can be evaluated at an arbitrary $\mathbf{x}\in\mathcal{X}$.
Bayesian Optimization(BO) algorithms have been deeply studied and used for HPO in many machine learning(ML) and deep neural network(DNN) problems. Recently, DNN HPO has gained greater importance with increasingly complex tasks that ML techniques cannot fully support.

\begin{figure*}[ht]
	\centering
	\begin{subfigure}[b]{.31\textwidth}
		\label{fig1:a}
		
		\includegraphics[width=\columnwidth]{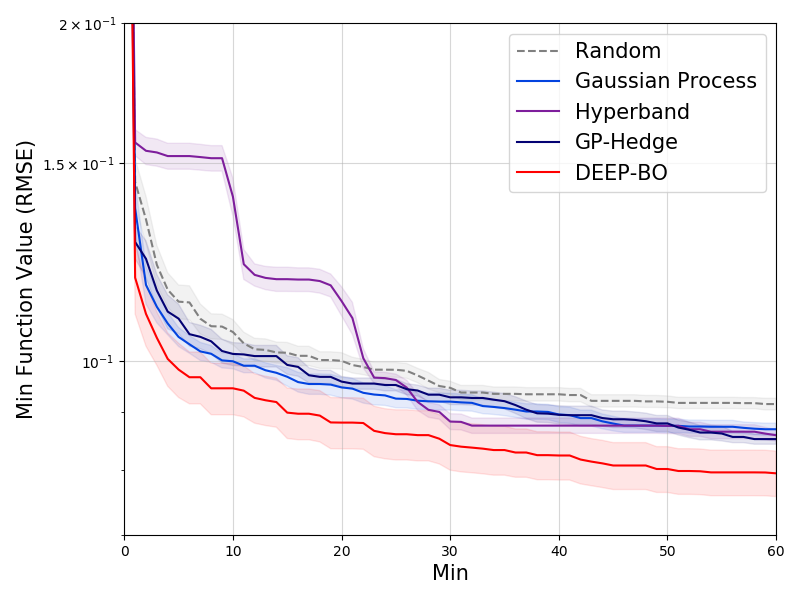}
		\caption{Regression}
	\end{subfigure}
	\begin{subfigure}[b]{.31\textwidth}
		\label{fig1:b}
		\includegraphics[width=\columnwidth]{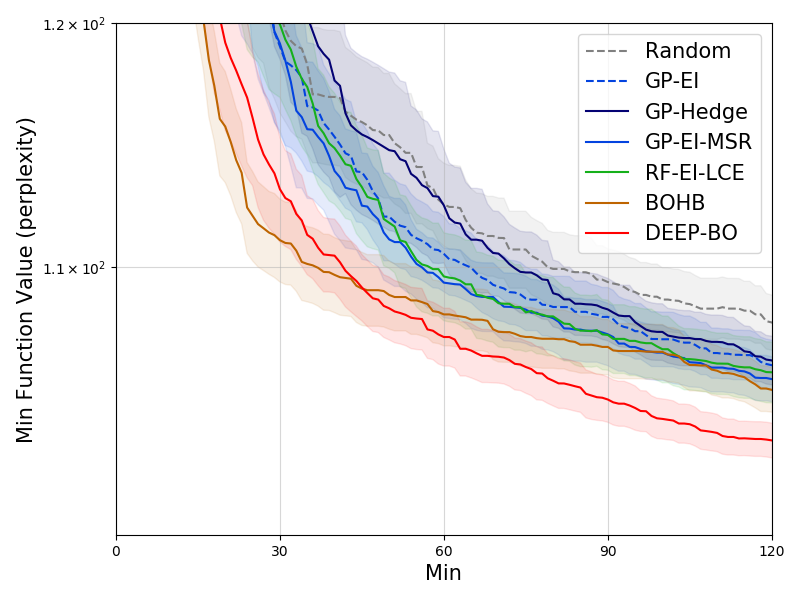}
		\caption{Language modeling}
	\end{subfigure}
	\begin{subfigure}[b]{.31\textwidth}
		\label{fig1:c}
		\includegraphics[width=\columnwidth]{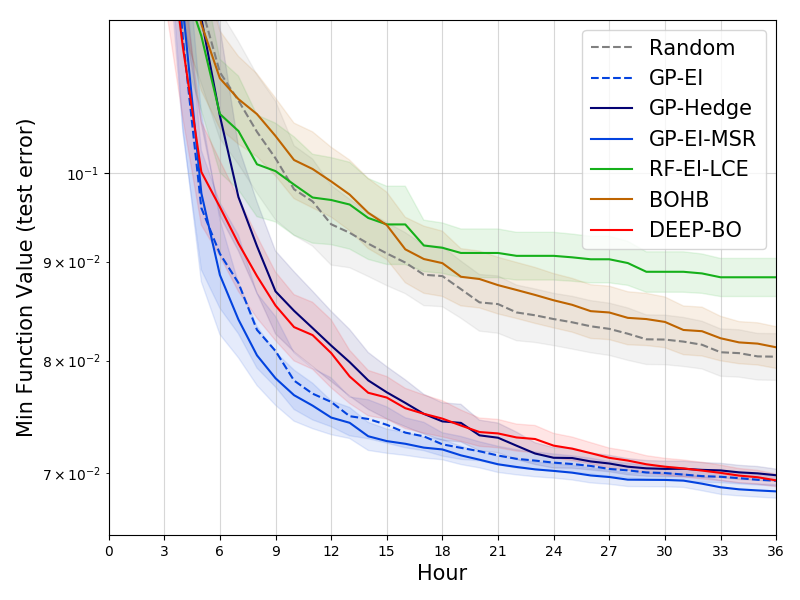} 
		\caption{Classification}
	\end{subfigure}

	\caption{
		HPO performance comparison on 3 different DNN tasks (lower is better):
		(a) MLP model on kin8nm. 
		(b) LSTM model on PTB, 
		(c) ResNet model on CIFAR-10. 
		Each plot compares min function value of different HPO algorithms as the optimization time $t$ increases.
		DEEP-BO is our method.		
		Note that the shaded areas are 0.25 $\sigma$ error.		
	}
\label{fig1}
\end{figure*}

We have observed that it is difficult to choose an appropriate HPO algorithm because the search space depends on the data characteristics and the DNN structure.
As shown in Figure~\ref{fig1}, we first evaluated 3 different DNN tasks using various HPO algorithms: random search~\citep{bergstra2012random}, Gaussian Process(GP-EI)~\citep{snoek2012practical}, an adaptive GP algorithm GP-Hedge~\citep{hoffman2011portfolio}, two speed-up BOs using Median Stopping Rule(GP-EI-MSR) and Learning Curve Extrapolation(RF-EI-LCE)~\citep{golovin2017google, domhan2015speeding}, bandit based approach Hyperband~\citep{li2017hyperband} and  BOHB~\citep{falkner2018bohb}.
No free lunch theorem proves that no algorithm can show superior performance in all cases~\citep{wolpert1997no}.
Nevertheless, DEEP-BO that this work introduces shows stable performance in most cases.

DEEP-BO (Diversified, Early-termination Enabled, Parallel BO) is an algorithm made more robust by strategically integrating 3 methods - BO diversification, safe early termination, and cost function transformation. We first detail how each method comprising DEEP-BO is developed to enhance its respective precedents and then proceed to observe how their combination outperforms existing strategies for DNN HPO. We discuss the individual strengths of the comprising methods and observe that their combined use in the form of DEEP-BO shows both greater performance and robustness than when used alone in enhancing traditional BO for DNN HPO tasks.

%% file: 2_related_works.tex
\section{Related Works}
\label{sec:related}


\subsection{Bayesian Optimization}
Bayesian optimization is an iterative process that sequentially chooses the next configuration to evaluate based on previous observations.
Three Sequential Model-Based Global Optimization (SMBO) algorithms, Spearmint~\citep{snoek2012practical}, SMAC~\citep{hutter2011sequential}, and TPE~\citep{bergstra2011algorithms}, were introduced and are compared in practical problems~\citep{eggensperger2013towards}. Acquisition functions such as Probability of Improvement (PI), Expected Improvement (EI), and Upper Confidence Bound (UCB) are widely used with these three BO algorithms.


\subsection{Diversification strategies}
Diversification strategy is introduced to minimize the risk of loss by splitting different categories of methods in various problems.
Diversification strategies have been successfully used in many fields such as communication~\citep{zheng2003diversity} and finance~\citep{markowitz1952portfolio} because of their theoretical and practical advantages.
Diversification strategies have also been applied to HPO of ML algorithms in various ways. 
GP-Hedge~\citep{hoffman2011portfolio} uses multiple acquisition functions together in online multi-armed bandit settings to improve HPO performance. For situations where there is no clear reason to choose a particular acquisition function, using a portfolio of them can be beneficial.

\subsection{Early stopping strategies}

Speed up practices today attempt to reap the benefits of HPO in the form of resource allocation strategies by considering learning dynamics of DNN.
One common prior is that learning curves typically increase and saturate. 
\cite{li2017hyperband} further articulates this using \textit{envelope functions}\footnote{It is defined as the maximum deviation of the intermediate losses from the terminal validation loss toward which the deviations are suggested to monotonically decrease with resources allocated.} before applying it to the strategy.
Even though asymptotic accuracy predictions vary from a heuristic rule~\citep{golovin2017google}, probabilistic modeling~\citep{domhan2015speeding},  to regression~\citep{baker2017accelerating,istrate2018tapas},  the termination criterion is set upon such an assumption.
Some works develop on these precedents. \cite{klein2016learning} uses Bayesian methods to ~\cite{domhan2015speeding}.
\cite{falkner2018bohb} applies BO to \cite{li2017hyperband} to capture the density of good configurations evaluated. Most of these works term their strategies as early stopping strategies.

\subsection{Parallel BO}

BO is a naturally sequential process, but parallelization can also be applied. 
Some studies address the following issues about parallel BO: scalability~\citep{springenberg2016bayesian}, batch configuration~\citep{wu2016parallel,nguyen2016budgeted}, and exploration-exploitation tradeoffs~\citep{desautels2014parallelizing}.
\cite{snoek2012practical} calculates Monte Carlo estimates of the acquisition function, which can be used in multiprocessor settings where the number of MCMC sampling can control the diversity of a GP model. This dynamic is briefly examined in this work.
Many other strategies have been introduced to increase the diversity of parallel BO algorithms: combining UCB and pure exploration~\citep{contal2013parallel}, using a penalized acquisition function~\citep{gonzalez2016batch}, introducing a randomized strategy to divide and conquer~\citep{wang2017batched} and modeling the diversity of a batch~\citep{kathuria2016batched}.

%% file: 3_deep_bo.tex
\section{DEEP-BO: An Enhanced Bayesian Algorithm for DNN HPO}
\label{sec:s-div}


We introduce three methods intended to enhance BO performance and robustness.
Appendix A details the full algorithm with more explanation.

\subsection{Diversification of BO}

While a typical BO algorithm uses a single modeling algorithm $\hat{f}$ for all iterations to approximate the true black-box function $f$, sequential diversification uses $N$ modeling algorithms ($\hat{f}_1,\dotsc,\hat{f}_N$) in sequence. 
This approach rotates over the ordered $N$ algorithms in sequence.
%
%
When $M$ parallel processors (workers) are available, the parallel diversification strategy can be used by utilizing $N$ modeling algorithms over $M$ processors. 
Assuming that the time required for evaluating $f(\mathbf{x}^*)$ is much longer than that required for modeling $\hat{f}_n$ using history $\mathcal{H}$, which is often true for heavy DNN problems, we can adopt a simple strategy where we assign the modeling result $\mathbf{x}^*=\operatorname*{arg\,max}_{\mathbf{x} \in \mathcal{X}} \hat{f}_n(\mathbf{x})$ to whichever worker that becomes available. 
Here, $n$ would be cycled from 1 to $N$ in sequence such that diversification can be enforced. 
%


\subsection{Early termination rule}

We introduce an early termination rule, which we name as the Compound Rule (CR), developed on the Median Stopping Rule (MSR)~\citep{golovin2017google} suggested to be robust.
We define \textit{Early Termination Rule} (ETR) as a rule that economizes on resources by prematurely terminating unpromising configurations~\footnote{Note that we avoided the term \textit{early stopping} because historically it has been used to refer to a specific process that prevents overfitting from excessive training ~\citep{prechelt1998automatic}.}. 
MSR records the mean performance up to some epoch $j$ for each configuration tried and uses its median value to evaluate the pending configuration's best performance until epoch $j$. Our ETR has 2 design parameters (max epoch $E$ and $\beta\in(0.0,0.5]$). $\beta=0.1$ is used here with less expected aggressiveness.
CR considers curves to be one of three types - highly untrainable, highly desirable, and the rest - and aims to find the first two with checkpoint-1 and checkpoint-2 respectively. Section~\ref{sec:discussion} and Appendix A and B contain further explanation this ETR. Below is a summary of the two checkpoints.

Checkpoint-1 discards highly untrainable configurations. It terminates the training of the current $k_{\text{th}}$ configuration $\mathbf{x}_k$ at epoch $j_1$ if its until-then best performance is strictly worse than the $\beta$ percentile value of the running averages $\bar{f}_{1:j_1}({\mathbf{x}})$ for all $\mathbf{x}\in{\mathcal{H}}$. The running average performance of $\mathbf{x}_k$ up to epoch $l$ is calculated as $\bar{f}_{1:l}({\mathbf{x}})=\frac{1}{l}\sum^l_{i=1}f_{i}({\mathbf{x}})$ where $f_{i}({\mathbf{x}})$ is performance of $\mathbf{x}$ at epoch $i$. Poor configurations consistently produce near random-guess performances and more likely arise when necessary priors for hyperparameters are unavailable which greatly limits the optimization potential~\citep{gulccehre2016knowledge}.
Checkpoint-1 exclusively aims to terminate them with a low ($\beta$ percentile; MSR uses 50th percentile) threshold.

Checkpoint-2 spares only highly desirable configurations that are likely to outperform the current best. 
$\mathbf{x}_k$ that survives checkpoint-1 is stopped at epoch $j_2$ if its until-then best performance is below the $(1-\beta)$ percentile value of the running averages $\bar{f}_{j_1:j_2}({\mathbf{x}})$ for $\mathbf{x}\in{S}$ where $S\subset{\mathcal{H}}$ for $\mathbf{x}$ that survived checkpoint-1. 
Envelope functions suggest more resources be needed to discern configuration quality since envelope overlaps decrease with resources more clearly.  
We observe this more evidently for highly desirable configurations. 
We thereby apply the intuition exclusively to discern these good configurations with $j_2$ proportional to threshold height. 
We also note that the observed range for checkpoint-2 starts from $j_1$, not the first epoch.


\subsection{Parallel BO}
%
%
%
As \cite{snoek2012practical} points out, the main problem with parallel workers is deciding what $\mathbf{x}$ to evaluate next while a set of points are still being evaluated. The simplest strategy is to ignore others and include only completely evaluated results into $\mathcal{H}$. This can result in duplicate selection where the chosen $\mathbf{x}^*$ is already being evaluated by another worker.
\cite{snoek2012practical} adds randomness through MCMC sampling to reduce duplicate selection. In the case of parallelization, using $N$ different algorithms naturally lowers the chance of duplicates because the algorithms tend to make different decisions even for the same $\mathcal{H}$. When $M$ is much larger than $N$, however, duplicate selection becomes more likely even for diversification. When it occurs, the following can be considered.  
\begin{itemize}
	\item \textit{na{\"i}ve}: evaluate the duplicate anyway
	\item \textit{random}: randomly select a new candidate
	\item \textit{next candidate}: choose the next best candidate that has not been selected before  
	\item \textit{in-progress}: temporarily add the premature evaluation results to $\mathcal{H}$ and find $\mathbf{x}^*$  
\end{itemize}
For DNN, \textit{in-progress} is attractive because even when the premature result is entirely wrong, it is self-corrected and replaced in $\mathcal{H}$ when evaluation completes. Experiments conducted with $N=M=6$ shows that \textit{in-progress} provides meaningful improvement while the others perform comparably.

Another important topic to consider is cost function transformation. 
For both improving modeling performance and robustness, we also introduce a \textit{hybrid transformation} which uses a linear scale when the error rate is above $\alpha$ and log transformation when the error rate is below $\alpha$.  
DNN is often used for classification tasks with the performance as an error rate. When the desired error rate is close to 0, it becomes difficult to accurately model $\hat{f}(\mathbf{x})$ using the original linear scale. 
For such case, applying log transformation, known to be effective for modeling nonlinear and multimodal functions~\citep{jones1998efficient}, can be advantageous. However, when the error rate is high, this log transformation can be harmful.  

%% file: 4_experiment.tex
\section{Empirical Results}

\begin{table}[b]
	
	\centering
	\caption{Benchmark problem settings.}
	\label{tab0}
	\resizebox{\columnwidth}{!}{
		\begin{threeparttable}
			\begin{tabular}{ccccccc}
				\toprule
				\multirow{2}{*}{\textbf{\begin{tabular}[c]{@{}c@{}}HPO Benchmark\end{tabular}}} & \multicolumn{3}{c}{\textbf{Number of hyperparameters}}                & \multirow{2}{*}{\textbf{\begin{tabular}[c]{@{}c@{}}Learning\\ rate scale\end{tabular}}} & \multirow{2}{*}{\textbf{\begin{tabular}[c]{@{}c@{}}Max epoch\end{tabular}}} &
				\multirow{2}{*}{\textbf{\begin{tabular}[c]{@{}c@{}}Number of\\ configurations\end{tabular}}}
				\\
				\cmidrule(l){2-4}
				& \textbf{discrete}           & \textbf{continuous}         & \textbf{categorical}        &                                      \\
				\midrule
				MNIST-LeNet1             & 6            & 3             & -              & log            & 15            & 20,000                   \\
				MNIST-LeNet2              & 3            & 3             & 3              & log             & 15            & 20,000               \\
				PTB-LSTM            & 3            & 5             & 1              & linear                & 15    & 20,000             \\
				CIFAR10-VGG            & 6            & 2             & 2              & log                & 50    & 20,000             \\
				CIFAR10-ResNet        & 2            & 4             & 1              & log                  & 100   & 7,000          \\
				CIFAR100-VGG            & 6            & 2             & 2              & log                & 50    & 20000           \\ 
				\bottomrule
			\end{tabular}
		\end{threeparttable}	
	}
\end{table}

In this study, we thoroughly evaluated DEEP-BO over 6 DNN HPO problems. 
As summarized in Table~\ref{tab0}, we chose the most popular benchmark datasets in the DNN community: MNIST, PTB(Penn Treebank), CIFAR-10, and CIFAR-100.
Three HPO tasks of CNN architectures (LeNet, VGG, ResNet) and one of RNN architecture (LSTM) are included.
Since practical model implementation almost always limits the number of hyperparameters human experts can adjust, seven to ten hyperparameters are selected for optimization for each benchmark task.
Learning rate, often the most critical, is included in all six problems.
To repeat the experiments effectively over multiple algorithms, 
we first created lookup tables for each task by pre-evaluating surrogate configurations sampled by Sobol sequences~\cite{sobol1967distribution}.
The appendix C specifies more details of each benchmark.

We use two different metrics to evaluate HPO algorithms. The first is the \textit{success rate}, the probability an HPO algorithm achieves a target goal within the given time budget $t$.
Assuming the algorithm takes time $\tau$, a random variable, to achieve a target accuracy $c$, the algorithm's \textit{success rate} at time $t$ can be defined as successfully finding $\hat{\mathbf{x}}$ such that $f(\hat{\mathbf{x}})>c$ before time $t$, or more simply as $\mathbf{P}(\tau \leq t)$. The second is \textit{expected time} to achieve a target accuracy expressed as $\mathbb{E}[\tau]$.


 
\begin{figure*}[!ht]
	\centering
	\begin{subfigure}[b]{.3\textwidth}
		\includegraphics[width=\columnwidth]{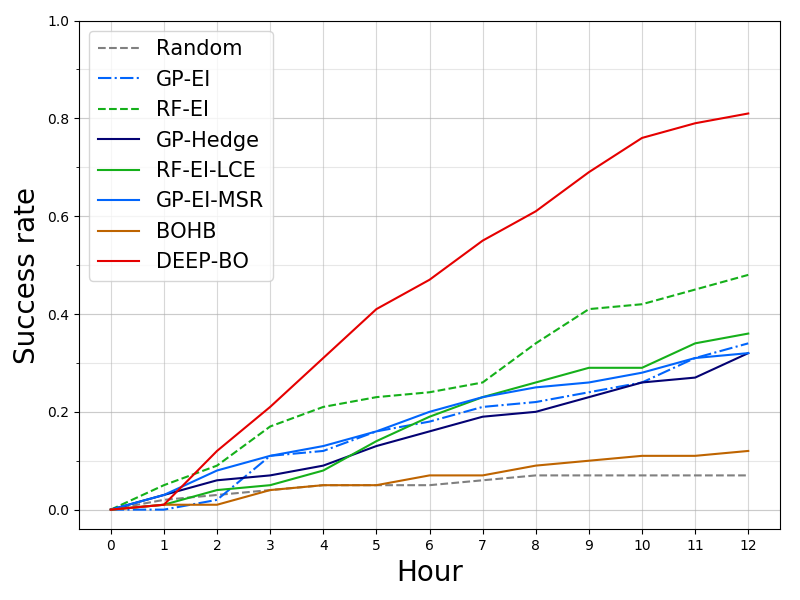}
		\caption{MNIST-LeNet1 ($M=1$)}
		\label{fig2:a}
	\end{subfigure}
	\begin{subfigure}[b]{.3\textwidth}
		\includegraphics[width=\columnwidth]{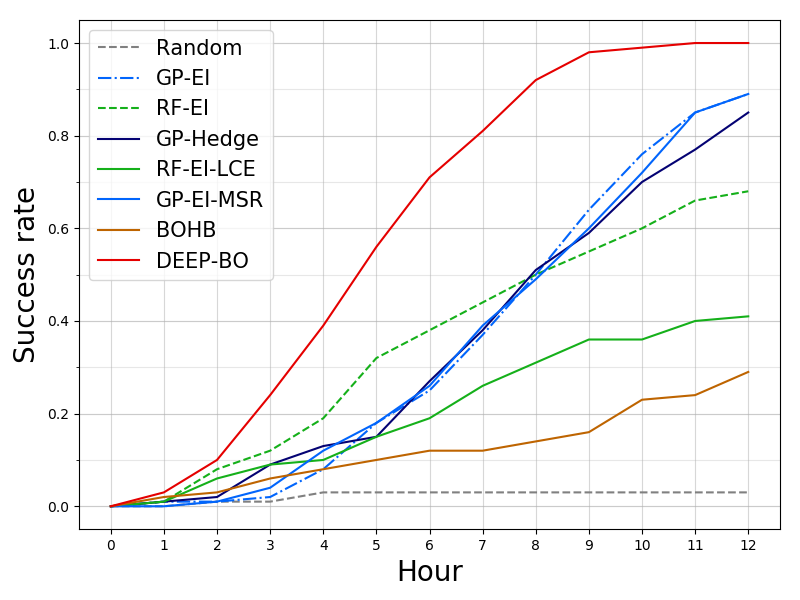}
		\caption{PTB-LSTM ($M=1$)}
		\label{fig2:b}
	\end{subfigure}
	\begin{subfigure}[b]{.3\textwidth}
		\includegraphics[width=\columnwidth]{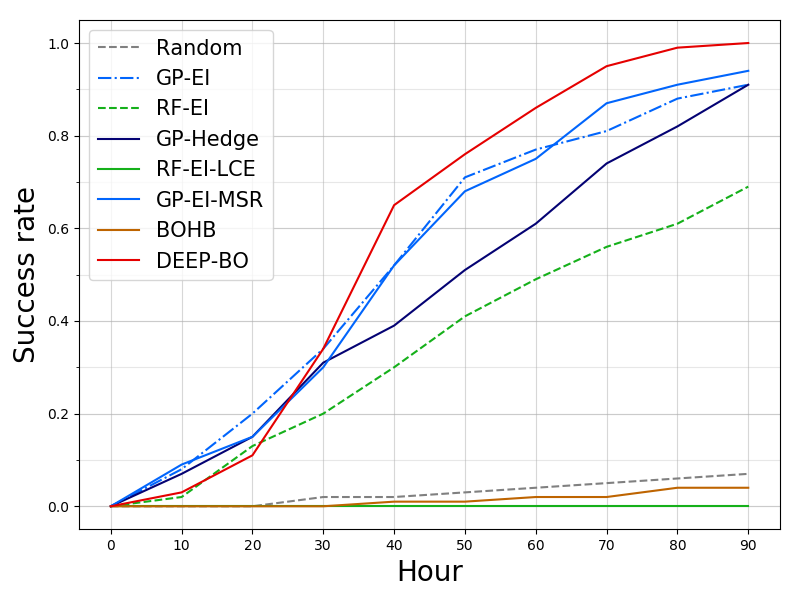}
		\caption{CIFAR10-ResNet ($M=1$)}
		\label{fig2:c}
	\end{subfigure}
	\begin{subfigure}[b]{.3\textwidth}
	\includegraphics[width=\columnwidth]{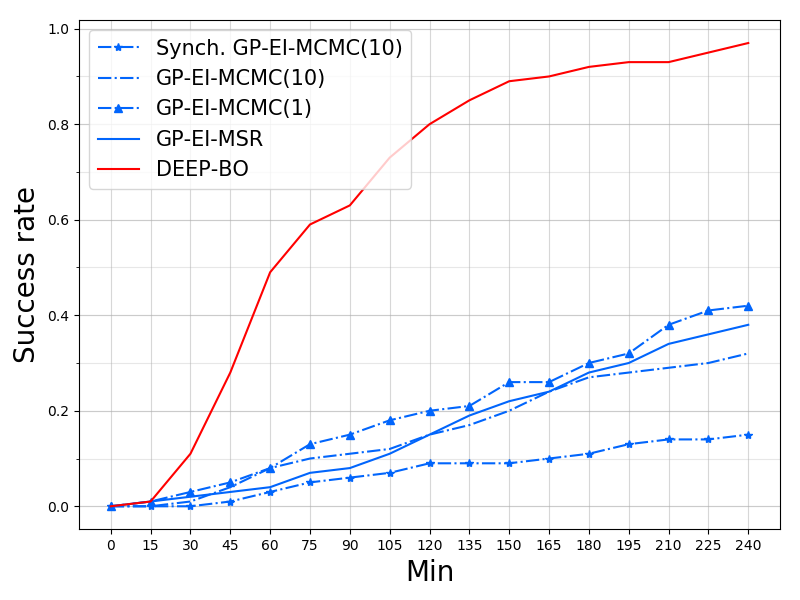}
	\caption{MNIST-LeNet1 ($M=6$)}
	\label{fig2:d}
	\end{subfigure}
	~ 
	\begin{subfigure}[b]{.3\textwidth}
		\includegraphics[width=\columnwidth]{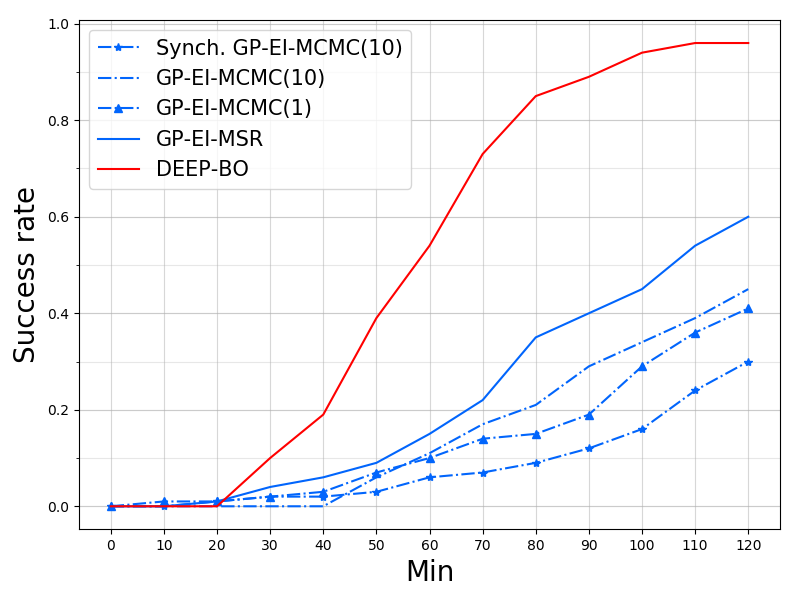}
		\caption{PTB-LSTM ($M=6$)}
		\label{fig2:e}
	\end{subfigure}
	~ 
	\begin{subfigure}[b]{.3\textwidth}
		\includegraphics[width=\columnwidth]{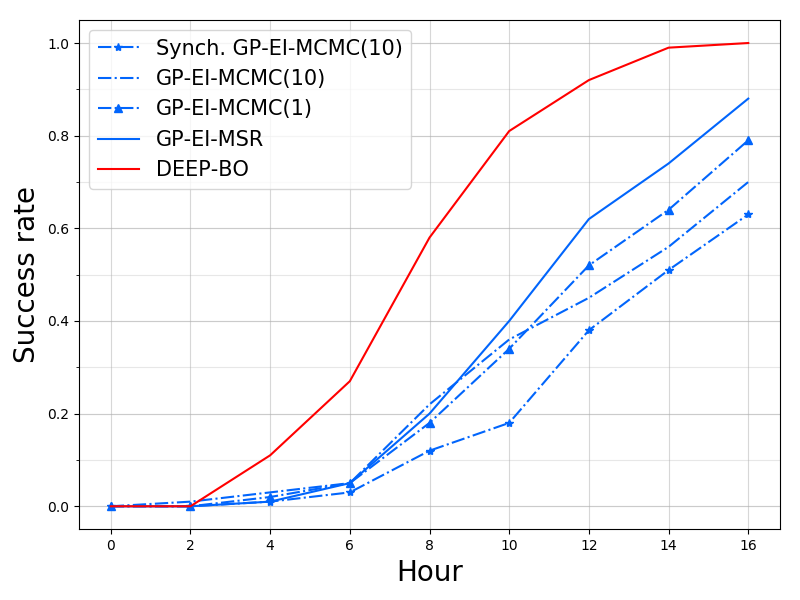}
		\caption{CIFAR10-ResNet ($M=6$)}
		\label{fig2:f}
	\end{subfigure}   

	\caption{
		Success rate to achieve a target error performance using $M$ processor(s).
		All experiments repeated the HPO run 100 times for each algorithm.
	}
	\label{fig2}
	
\end{figure*}

Figure~~\ref{fig2} shows the \textit{success rate} plots of DEEP-BO and other algorithms in 3 of 6 benchmark cases. 
Unlike other benchmark algorithms, DEEP-BO consistently shows top-level performance for all 3 benchmark cases shown. 
Appendix D of supplementary materials details plots for all benchmarks.

Even under the conditions of using a single processor, DEEP-BO is not only robust in all benchmarks, but also has excellent performance.
In Figure~~\ref{fig2:a}$\sim$\ref{fig2:c}, we evaluated our method using $M$ = 1, $N$ = 6, $\alpha$ = 0.3, $\beta$ = 0.1 and compared it  with an adaptive algorithm GP-Hedge, two speed up algorithms (MSR, LCE) and BOHB, a state-of-the-art algorithm by \cite{falkner2018bohb}.

The performance of DEEP-BO under multiprocessing conditions is impressive.
Figure~\ref{fig2:d}$\sim$\ref{fig2:f} plot \textit{success rates} evaluated in 6 parallel processors($N$ = 6, $M$ = 6, $\alpha$ = 0.3, $\beta$ = 0.1). 
DEEP-BO is compared with four different settings of GP-EI algorithms regarding synchronization, MCMC sampling, and early termination.  
In GP-EI algorithms, MCMC sampling explained in \cite{snoek2012practical} was adopted with either 10 or 1 samples used. 
The GP-EI-MCMC(10) was the baseline algorithm, and GP-EI-MCMC(1) used only one sample to increase randomness. 
`Synch,GP-EI-MCMC(10)' was a special setting where all the six processors were synchronized for modeling such that there is no difficulty with handling incomplete evaluations. 
All of six processors started simultaneously and had to wait until the evaluations of all six completed such that the next top 6 candidates can be launched together. 
It performs poorly as expected, but is provided as a benchmark. 
Both GP-EI-MCMC(1) and GP-EI-MCMC(10) observably performed comparably and the increased randomness did not a prompt distinct improvement. 
MSR enabled GP-EI tends to increase the performance but, it is not significantly improved.

\begin{table*}[!ht]
	\centering

	\caption{Summary of performance achieving top-10 accuracy. 
		For \textit{success rate}, the numbers in parenthesis of each benchmark indicate the checkpoints when the number of processors is 1 and 6, respectively.
		We only show the competitors' performance due to the page limit.
		See Appendix D for full benchmark results.} 

	\label{tab-result}
	\resizebox{\columnwidth}{!}{
	\begin{tabular}{l|l|c|c|c|c|c|c|c}
		\toprule
		\multirow{2}{*}{\textbf{Measure}} & \multicolumn{1}{l|}{\multirow{2}{*}{\textbf{Benchmark}}} & \multicolumn{4}{c|}{Single processor (\boldmath$M = 1$)} & \multicolumn{3}{c}{ Six processors (\boldmath$M = 6$)} \\
		\cmidrule(l){3-9}
		& \multicolumn{1}{l|}{} & \textbf{GP-Hedge} & \textbf{GP-EI-MSR} & \textbf{BOHB} & \textbf{DEEP-BO} & \textbf{GP-EI} & \textbf{GP-EI-MSR} & \textbf{DEEP-BO} \\
		\midrule

		\multirow{7}{*}{\begin{tabular}[c]{@{}l@{}}Success\\    rate \end{tabular}} & MNIST-LeNet1 \space\space\space\space\space(12h / 2h) & 32\% & 32\% & 12\% & \textbf{81\%} & 15\% & 15\% & \textbf{80\%} \\		
		& MNIST-LeNet2 \space\space\space\space\space(12h / 2h) & 39\% & 48\% & 1\% & \textbf{82\%} & 39\% & 41\% & \textbf{95\%} \\
		& PTB-LSTM \space\space\space\space\space\space\space\space\space\space\space(12h / 2h) & 85\% & 89\% & 29\% & \textbf{100\%} & 45\% & 60\% & \textbf{96\%} \\
		& CIFAR10-VGG \space\space\space\space\space\space\space(6h / 1h) & 53\% & 48\% & 0\% & \textbf{80\%} & 32\% & 50\% & \textbf{75\%} \\
		& CIFAR10-ResNet (90h / 15h) & 91\% & 94\% & 4\% & \textbf{100\%} & 61\% & 81\% & \textbf{99\%} \\
		& CIFAR100-VGG \space\space\space\space\space(6h / 1h) & 92\% & 85\% & 4\% & \textbf{100\%} & 30\% & 43\% & \textbf{99\%} \\
		\cmidrule(l){2-9}
		& Mean & 66\% & 66\% & 8\% & \textbf{91\%} & 37\% & 48\% & \textbf{91\%} \\
		\midrule
		\multirow{6}{*}{\begin{tabular}[c]{@{}l@{}}Expected\\time  \\(hour)\end{tabular}} & MNIST-LeNet1 & $17.0\pm8.2$ & $16.3\pm8.6$ & $20.7\pm6.4$ & \boldmath$8.2\pm8.4$ & $10.4\pm8.5$ & $4.0\pm1.5$ & \boldmath$1.4\pm1.0$ \\

		& MNIST-LeNet2 & $29.3\pm31.2$ & $24.2\pm30.4$ & $35.8\pm12.1$ & \boldmath$7.6\pm10.4$ & $4.6\pm5.4$ & $2.4\pm1.4$ & \boldmath$0.8\pm1.0$ \\
		& PTB-LSTM & $8.3\pm3.6$ & $7.9\pm2.8$ & $28.3\pm22.7$ & \boldmath$4.7\pm2.2$ & $2.1\pm0.7$ & $1.8\pm0.8$ & \boldmath$1.0\pm0.4$ \\
		& CIFAR10-VGG & $6.6\pm4.0$ & $6.8\pm4.3$ & $29.4\pm1.8$ & \boldmath$3.4\pm2.5$ & $1.5\pm0.9$ & $1.3\pm0.8$ & \boldmath$0.8\pm0.4$ \\
		& CIFAR10-ResNet & $50.6\pm29.6$ & $43.8\pm25.4$ & $129.8\pm17.8$ & \boldmath$38.4\pm17.3$ & $13.0\pm5.5$ & $11.4\pm4.1$ & \boldmath$7.7\pm2.5$ \\
		& CIFAR100-VGG & $4.1\pm2.1$ & $3.5\pm1.5$ & $24.8\pm9.9$ & \boldmath$1.9\pm0.9$ & $1.2\pm0.5$ & $1.1\pm0.4$ & \boldmath$0.5\pm0.2$ \\
		\cmidrule(l){2-9}
		& Mean & $25.7$ & $17.2$ & $44.8$ & \boldmath$10.7$ & $5.5$ & 3.7\% & \boldmath$2.0$ \\		
		\bottomrule
	\end{tabular}
}
\end{table*}

DEEP-BO notably enhances performance with a single processor. DEEP-BO with 6 processors can enhance more through parallelization.
As shown in Figure~\ref{tab-result}, DEEP-BO outperforms in all benchmarks and seems robust than competitors.
In terms of \textit{expected time}, DEEP-BO performance using 1 processor is 10.7 hours, but that with 6 processors is only 2 hours (5.35 times faster). 
The state of the art algorithm, BOHB, did not perform well in these benchmarks even though we had tried to set its hyperparameters properly. 
We discuss why in the section~\ref{sec:discussion}.


\begin{table*}[!ht]
	\centering
	\caption{Ablation test results with DEEP-BO using 6 processors. \textbf{Bold} refers to the best performance and \textit{Italic} refers to comparable results.} 
	\label{tab-ablation}
	\resizebox{\columnwidth}{!}{

		\begin{tabular}{lllcccccc}
			\toprule
			\multirow{2}{*}{\begin{tabular}[c]{@{}l@{}}\textbf{Hybrid}\\  \textbf{Transformation}\end{tabular}} & \multirow{2}{*}{\begin{tabular}[c]{@{}l@{}}\textbf{Early}\\   \textbf{Termination}\end{tabular}} & \multirow{2}{*}{\begin{tabular}[c]{@{}l@{}}\textbf{In-}\\ \textbf{progress}\end{tabular}} & \multicolumn{6}{c}{\textbf{Expected time (hour)}}                                                                                                                                                                             \\
			\cmidrule(l){4-9}
			&                                    &                                                                                               & \multicolumn{1}{c}{\begin{tabular}[c]{@{}l@{}}MNIST-\\LeNet1\end{tabular}} & \multicolumn{1}{c}{\begin{tabular}[c]{@{}l@{}}MNIST-\\LeNet2\end{tabular}} & \multicolumn{1}{c}{\begin{tabular}[c]{@{}l@{}}PTB-\\LSTM\end{tabular}} & \multicolumn{1}{c}{\begin{tabular}[c]{@{}l@{}}CIFAR10-\\VGG\end{tabular}} & \multicolumn{1}{c}{\begin{tabular}[c]{@{}l@{}}CIFAR10-\\ResNet\end{tabular}} & \multicolumn{1}{c}{\begin{tabular}[c]{@{}l@{}}CIFAR100-\\VGG\end{tabular}} \\
			\midrule
			on                                                                                    & on                                & on                                                                                   & $\mathit{1.4\pm1.0}$                               & $\mathit{0.8\pm1.0}$                               & \boldmath$1.0\pm0.4$                           & $\mathit{0.8\pm0.4}$                              & \boldmath$7.7\pm2.5$                                 & \boldmath$0.5\pm0.2$                               \\
			on                                                                                    & on                                & off                                                                                         & $1.6\pm1.0$                               & \boldmath$0.7\pm0.5$                               & $\mathit{1.1\pm0.5}$                           & $0.9\pm0.6$                              & $10.5\pm3.8$                                 & $0.7\pm0.3$                               \\			
			on                                                                                    & off                                 & on                                                                                   & \boldmath$1.3\pm0.7$                               & $1.0\pm1.5$                               & $1.2\pm0.5$                           & $0.9\pm0.6$                              & $\mathit{8.6\pm2.9}$                                 & $0.8\pm0.3$                               \\
			on                                                                                    & off                                 & off                                                                                         & $1.8\pm1.0$                               & $1.4\pm1.5$                               & $1.4\pm0.6$                           & $1.4\pm0.8$                              & $11.6\pm4.1$                                 & $1.1\pm0.3$                               \\

			off                                                                                        & on                                & on                                                                                   & $2.2\pm1.7$                               & $1.0\pm0.8$                               & $1.5\pm0.7$                           & \boldmath$0.7\pm0.5$                              & $8.9\pm3.3$                                 & \boldmath$0.5\pm0.2$                               \\

			off                                                                                        & on                                & off                                                                                         & $2.7\pm2.0$                               & $1.2\pm1.0$                               & $1.7\pm0.8$                           & $1.0\pm0.7$                              & $10.6\pm4.4$                                 & $0.8\pm0.3$                               \\

			off                                                                                        & off                                 & on                                                                                   & $2.4\pm1.8$                               & $1.5\pm1.1$                               & $1.8\pm0.7$                           & $1.1\pm0.7$                              & $10.5\pm4.4$                                 & $0.8\pm0.3$                               \\

			off                                                                                        & off                                 & off                                                                                         & $2.4\pm1.6$                               & $1.8\pm1.3$                               & $2.1\pm0.9$                           & $1.2\pm0.8$                              & $12.7\pm4.9$                                 & $1.1\pm0.4$   \\
			
			\bottomrule                           
		\end{tabular}
	}
\end{table*}
The ablation study in Table~\ref{tab-ablation} shows that, under all conditions, components applied to DEEP-BO do not always perform well but are generally stable.
Table~\ref{tab-ablation} shows (on, on, on) produces the most successful performances for 3 of 6 tasks and second best performances at worst.
Using this with some components turned off can be a little harmful, but it is shown that all the best cases were where at least 2 of the 3 switches were on.
Overall, (on, on, on) is a beneficial strategy for all benchmarks.

%% file: 5_discussion.tex
\section{Discussion}
\label{sec:discussion}
A few of the core design principles of DEEP-BO are explained and discussed in this section. 

\subsection{Diversity gain and cooperation effect}


\begin{figure}
	\centering
	\begin{minipage}{.30\textwidth}
		\centering
		\includegraphics[width=\columnwidth]{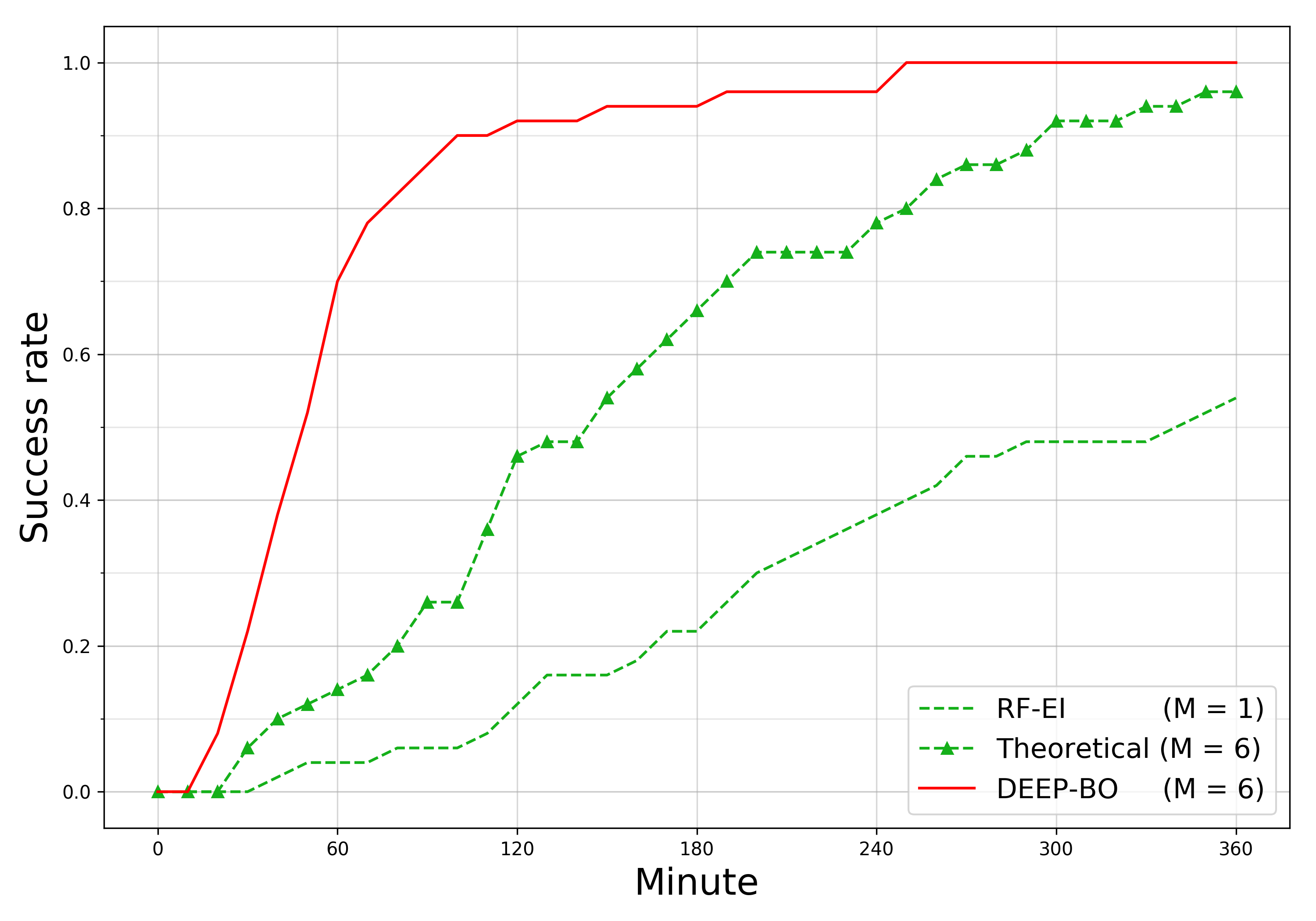}
		\caption{Diversity gain}
		\label{fig3}
	\end{minipage}%
	\begin{minipage}{.70\textwidth}
		\centering
			\begin{subfigure}[b]{.44\textwidth}
			\includegraphics[width=\columnwidth]{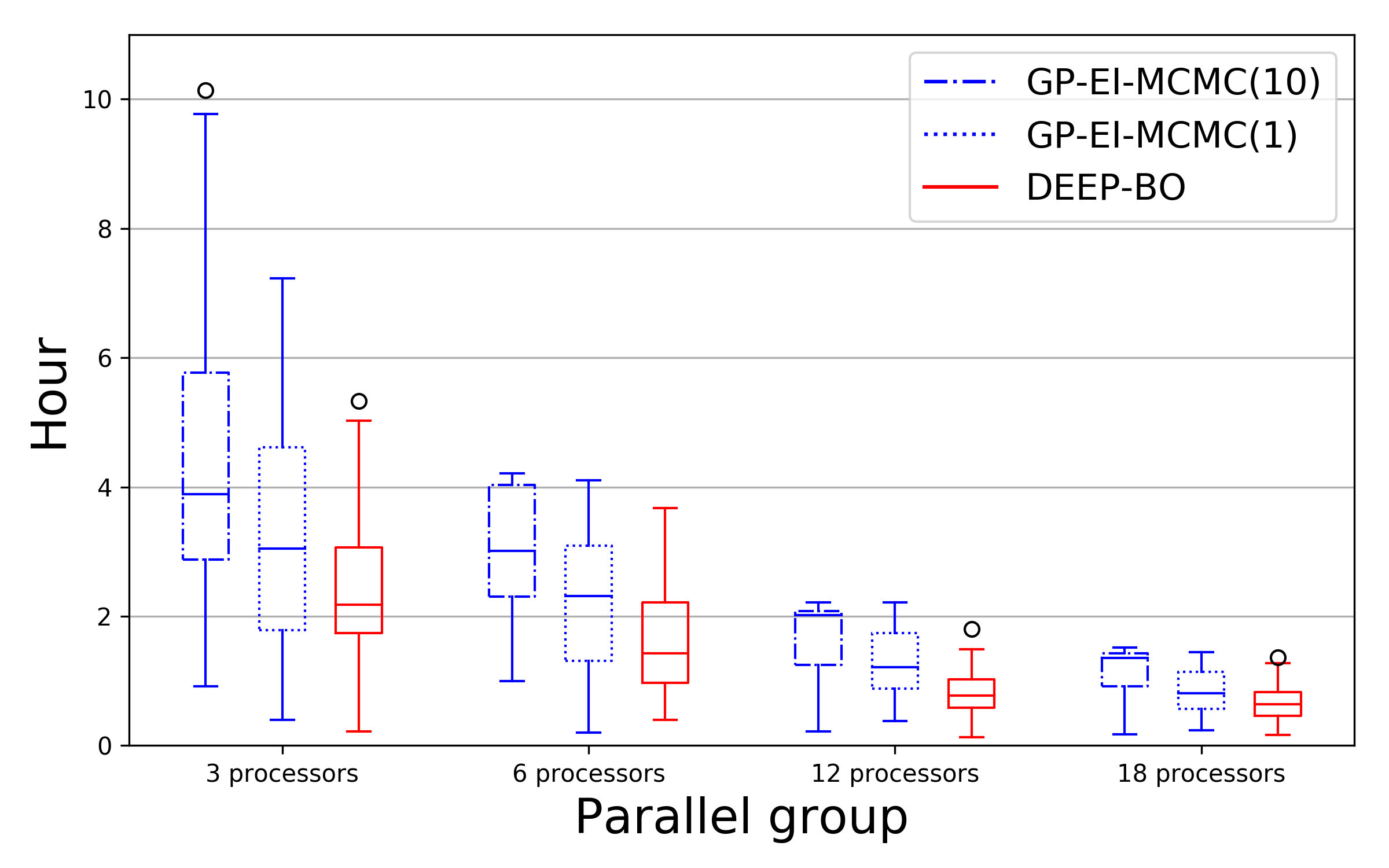}
			\caption{Expected time}
			\label{fig4:a}
		\end{subfigure}
		\begin{subfigure}[b]{.45\textwidth}
			\includegraphics[width=\columnwidth]{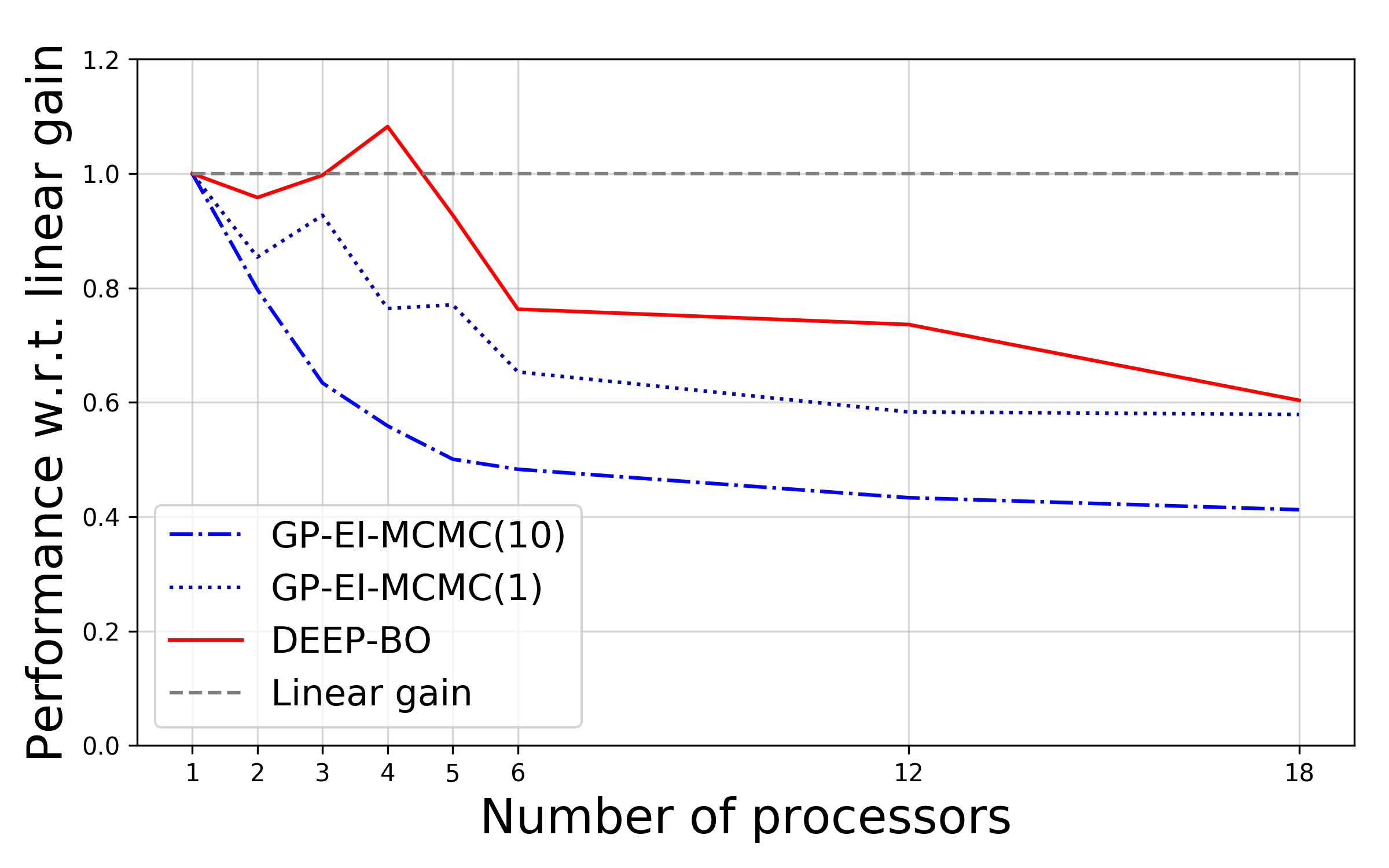}
			\caption{Performance w.r.t. linear gain}
			\label{fig4:b}
		\end{subfigure}
		\caption{Parallelization loss.}
		\label{fig4}
	\end{minipage}
\end{figure}

For an HPO algorithm, we can define its failure rate as $p_f = 1-\mathbf{P}(\tau \leq t)$ where $\tau$ is a random variable representing the algorithm's time to achieve the target accuracy $c$.
If the algorithm is run over $M$ parallel workers without sharing their histories $\{\mathcal{H}_m\}$ among the $M$ workers, the chance that all $M$ attempts fail is simply $p_f^M$ because $\tau$ forms an i.i.d. process. 
Therefore, we can see that $M^{\text{th}}$ order diversity gain (increase in the exponent of the failure rate) is achieved at the cost of $M$ times increase in resources. 
In Figure~\ref{fig3}, success rate curves are shown for a benchmark problem. RF-EI($M$=1) is the baseline with success rate equal to $1-p_f$, and Theoretical($M$=6) is a synthetic curve calculated from $1-p_f^6$. 
Clearly, success rate of Theoretical($M$=6) ramps up much faster than DEEP-BO($M$=1) and demonstrates the power of order six diversity. 
The exciting part is the performance of DEEP-BO($M$=6) that is shown together. DEEP-BO($M$=6) overwhelmingly outperforms Theoretical($M$=6), indicating that it can achieve much more than order six diversity ($1-p_f^6$). 

A large portion of the substantial extra improvement is due to the mix of $N=6$ individual models with different modeling characteristics,
and the effect comes in two-fold. The first is the different (ideally independent)
characteristics of the $N$ models where at least one might work well. This is 
another type of diversity that is different from the $p_f^M$ effect, and basically says 
at least one of $p_{f,1}, \dots, p_{f,N}$ will likely be small over the $N$ individual models 
for any given task. 
The second is the \textit{cooperative} nature of the diversified algorithm thanks to the sharing of $\{\mathcal{H}_n\}$. 
When an individual model \textit{A} is not capable of selecting good target points for exploration and is stuck in a bad region,
some of the other models might make better selections 
that serve well as exploratory points to the model \textit{A}. Then the poorly performing 
model \textit{A} can escape from the bad cycle and start to perform well. In our empirical experiments, we extensively tried 
multi-armed-bandit-like approaches to provide more turns to the better performing individual models, but the strategy actually turned 
out to hurt the overall performance. It turns out that it is better to give a fair chance to even the worst performing 
model such that its different modeling characteristic can be utilized. Of course, this is assuming that even the worst performing model 
is generally a plausible and useful model for DNN HPO.

\subsection{Parallelization loss}
\label{sec:loss}
When the \textit{expected time} is considered, it would be ideal if the time is reduced by $1/M$ by using $M$ processors. We examined if such a linear gain in performance can be retained as $M$ becomes large. In Figure~\ref{fig4:a}, distributions of the expected time are shown for 3, 6, 12, and 18 processors. Expected time and variance can be seen to decrease as the number of processor increases. As for GP-EI, we have found that MCMC(1) performs better probably because the increased randomness is helpful when multiple processors are used. 
In Figure~\ref{fig4:b}, the expected time performance with respect to $\times M$ is shown after normalization. The dotted gray line shows the $\times M$ performance. All the algorithms show the decrease in the normalized performance as $M$ is increased. But the values do not drop very fast after $M$ reaches 4$\sim$5 processors. For parallel diversification, interestingly the performance even went above linear gain when $M$ is between 2 and 4. This is due to the diversification and cooperation. 


\subsection{Robust termination criterion}
\label{sec:etr}

\begin{figure*}[ht]
	\centering
	\begin{subfigure}[b]{.32\textwidth}
		\includegraphics[width=\columnwidth]{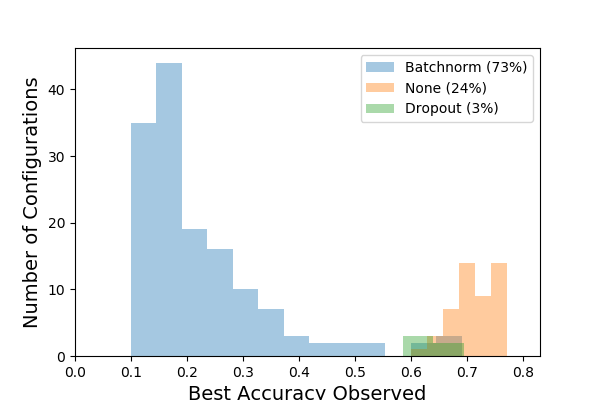}
		\caption{Initial performance}
		\label{fig2-1:a}
	\end{subfigure}
	\begin{subfigure}[b]{.32\textwidth}
		\includegraphics[width=\columnwidth]{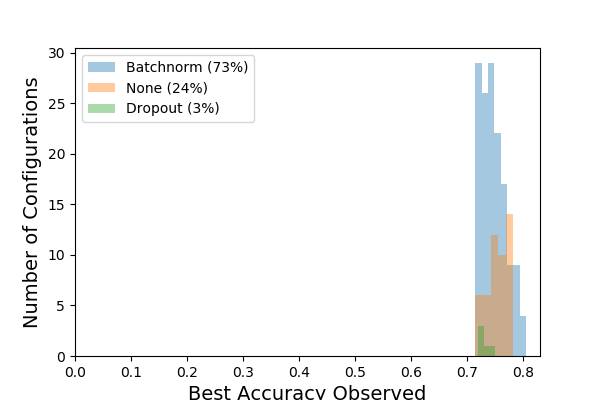}
		\caption{Later performance}
		\label{fig2-1:b}
	\end{subfigure}
	\begin{subfigure}[b]{.32\textwidth}
		\includegraphics[width=\columnwidth]{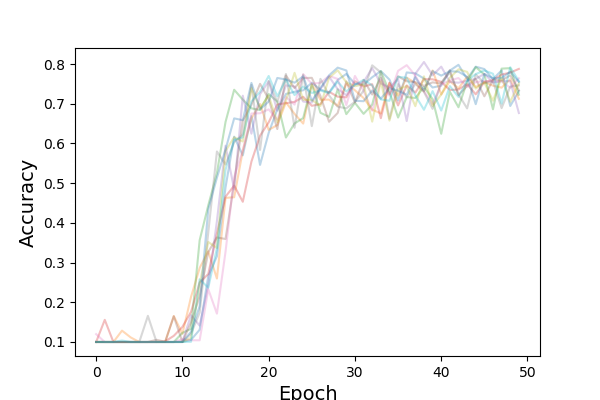}
		\caption{Top-10 performing curves}
		\label{fig2-1:c}
	\end{subfigure}
	\caption{These are based on the pre-evaluated CIFAR10-VGG surrogate configurations which included one of the three: batchnorm, dropout, or none. \ref{fig2-1:a} and \ref{fig2-1:b} show the performance distribution these 3 groups for top-200 curves. Configurations with batchnorm initially (epoch $[0,0.25E]$) performs worst but later (epoch $[0.75E,E]$) performs much better, signifying that early epoch information may be misleading. This relates to conjectures from \cite{choi2018difficulty}. In fact, all top-10 curves (\ref{fig2-1:c}) were from configurations that had batchnorm whose group initially performs worst.}
	\label{fig2-1}
\end{figure*}

Evaluating configurations very early can much save resources, but is possibly hazardous for BO.
CR, our ETR, intentionally does not locate $j_1$ in early epochs considering cases illustrated in Figure~\ref{fig2-1} for configurations with "batchnorm" whose termination can prompt false alarms that harm BO. 
Checkpoint-1($j_1$) here was set as rounded $E/2$ to lessen this possibility. 
Similar reasoning goes for checkpoint-2 whose observation start point for the scope of epochs observed is set as epoch $j_1$, not the first epoch. 
With all other conditions fixed for former-half epochs, the inter-dataset $\mu$ and $\sigma$ of \textit{survivor rank regret} is strictly counter-proportional to the observation start point for checkpoint-2 of our ETR (see Appendix B for details). 
These correlations strongly indicate that a delayed observation start point may encompass more reliable information (smaller $\mu$) and allow greater robustness (smaller $\sigma$).

Notable observations with our ETR are DEEP-BO's results compared to MSR and BOHB in Figure~\ref{fig2} and the comparison of ablation results about "Early Termination" in Table~\ref{tab-ablation}. That DEEP-BO outperforms MSR in our benchmarks is not too surprising as our rule was developed on MSR. The more interesting part is that from the perspective of \textit{success rates}, DEEP-BO outperforms BOHB even though both methods conceptually refer to similar parts of the \textit{envelope function}~\citep{li2017hyperband}.

DEEP-BO abstains from evaluating too early and specifies \textit{envelope functions} to be more evidently applicable to well-trainable configurations. BOHB evaluates from early epochs which can be dangerous for cases like Figure~\ref{fig2-1} particularly when all observed best configurations have poor initial performances. Terminating very early does help BO's explorability and allows the algorithm to reach easy goal performances very rapidly. However, this could increase false alarms, hindering the algorithm from performing beyond such easy targets. \textit{Success rate} is the average of Boolean results where the iteration is matched with 0 if it does not exceed goal performance. The goal for our study is set high and thereby BOHB could have failed drastically. 

\begin{figure*}[ht]
	\centering
	\begin{subfigure}[t]{.28\textwidth}
		\includegraphics[width=\textwidth]{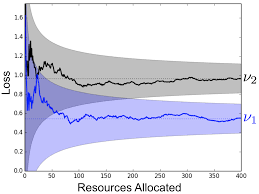}
		\caption{Envelope Function Concept}
		\label{fig6-2:a}
	\end{subfigure}
	\begin{subfigure}[t]{.35\textwidth}
		\includegraphics[width=\textwidth]{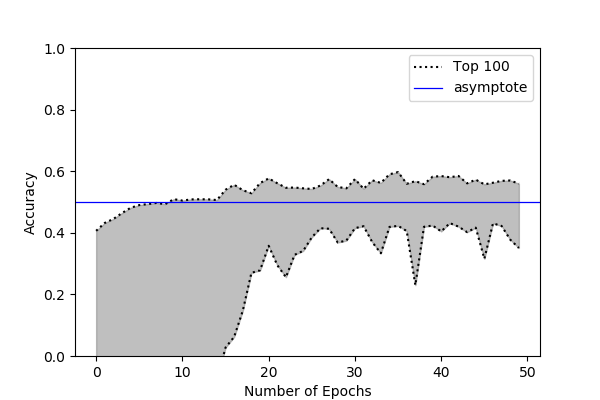}
		\caption{CIFAR10-VGG's Envelopes}
		\label{fig6-2:b}
	\end{subfigure}
	\begin{subfigure}[t]{.35\textwidth}
		\includegraphics[width=\textwidth]{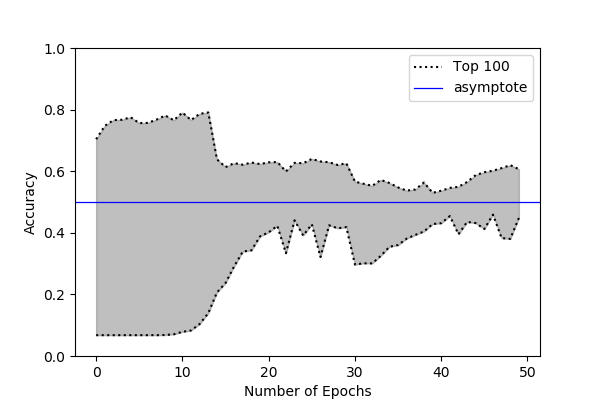}
		\caption{CIFAR100-VGG's Envelopes}
		\label{fig6-2:c}
	\end{subfigure}
	\caption{\textit{Envelope function} expectations~\citep{li2017hyperband} and their actual observations in CIFAR10/100-VGG for top-100 curves in total. Asymptotes (terminal performances) are approximated by the mean of the latter half performances and aligned at 0.5 for comparison.}
	\label{fig5}
\end{figure*}

\textit{Envelope function} overlappings should be minimal to be applicable. 
Though this is more evidently observed with high performing configurations, sporadic fluctuations that can bother evaluation still exist as Figure~\ref{fig5} illustrates. 
Thereby, it would be crucial for the ETR to be less fluctuation-prone. 
This can be aided by referring to a range of epochs rather than just a single epoch. 
Our rule and MSR are designed in such a way and therefore are in advantage in regards of robustness.

Table~\ref{tab-ablation} shows both the strength of DEEP-BO's ETR and the general limit of ETRs. 
The fact that DEEP-BO with CR was best for 5 out of 6 cases heuristically supports its robustness. 
However, one case revealed that there might be no always good ETR. 
CR was designed to be more robust than its precedents but still has its limits when used alone. 
Observations recommend ETRs be used along with other methods that help BO's robustness as this work attempts.

%% file: 6_conclusion.tex
\section{Conclusion}
\label{sec:conclusion}

In this work, we proposed three different strategies to enhance Bayesian Optimization for DNN HPO with a pursuit of improved average performance and enhanced robustness. 
We first adopted the concept of diversification. Individual BO models tend to focus on different parts of search space due to the different modeling characteristics. With diversification, each algorithm ends up serving as an excellent exploration guide to the other individual algorithms while being a greedy exploiter for its own selection. 
Then, we proposed an early termination strategy that works well by being conservative. Learning curves of DNN
can show distinct patterns depending on what options (e.g. batchnorm) are turned on, and it is very important to avoid being too aggressive. 
%
For parallel processing, we suggested the \textit{in-progress} method that uses premature evaluation results of pending DNN evaluations. Even when a premature evaluation is inaccurate and harmful, the problem is self-corrected after completing the evaluation. 
Additionally, we introduced a cost function transformation technique that can be helpful.
Overall, the resulting DEEP-BO works very well and outperformed the existing state-of-the-art algorithms
for the six benchmark DNN HPO tasks that we have investigated. 
%

%% file: supple-a.tex
\section*{Appendix A.\quad Algorithm Description}


\begin{algorithm}[H]
	\caption*{Algorithm A1. DEEP-BO}
	\label{alg:DEEP-BO}
	\begin{algorithmic}
		\State {\bfseries Inputs:} DNN $f$, modeling algorithm $\hat{f}_1$, $\dotsc$, $\hat{f}_N$, processor $p_1$, $\dotsc$, $p_M$, hyperparameter space $\mathcal{X}$, target accuracy $c$, cost function transformation $g$, early termination functions $h$ and $l$, hybrid transformation parameter $\alpha$, early termination parameter $\beta$, max epoch $E$.
		\State 
		\State $\mathcal{H}\leftarrow\emptyset$ \ \ \ \ \ \ \ \ \ \ \ \ \ \ \ \ \ \ \ \ \ \ \ \ \ \ \ \ \ \ \ \ \ \ \ \ \ \ \ \ \ \ \ \ \ \ \ \ \ \ \ \ \ \ \ \ \ \ \ \ \ \ \ \ \ \ \ \ \ \ \ \ \ \ \ \ \ \ \ \ \ \ \ \ \ \ \ \ \ \ \ \ \ \ \ \ \ \ \ \ \ \ \ \ \ \ \ \ \ \ \ \# \textit{global shared history}
		\For{$i=1, 2, \dotsc$}
		\State $n \leftarrow \mod(i, N) + 1$
		\State $m \leftarrow \mod(i, M) + 1$  

		\State Update $\hat{f}_n$ with $\mathcal{H}$ on $p_m$	\ \ \ \ \ \ \ \ \ \ \ \ \ \ \ \ \ \ \ \ \ \ \ \ \ \ \ \ \ \ \ \ \ \ \ \ \ \ \ \ \ \ \ \ \ \ \ \ \ \ \ \ \# \textit{preset $\hat{f}_n$ to $p_m$ when $N == M$}		
		\State Select $\mathbf{x}^* \in \operatorname*{arg\,max}_{\mathbf{x} \in \mathcal{X}} \hat{f}_n(\mathbf{x})$ on $p_m$
		\State $\mathcal{A}\leftarrow\emptyset$, ${y}^{*} \leftarrow 0$
		\For{$j=1, \dotsc,  E$}	\ \ \ \ \ \ \ \ \ \ \ \ \ \ \ \ \ \ \ \ \ \ \ \ \ \ \ \ \ \ \ \ \ \ \ \ \ \ \ \ \ \ \ \ \ \ \ \ \ \ \ \ \ \ \ \ \ \ \ \ \ \ \ \ \ \ \ \ \ \ \ \ \ \ \ \# \textit{execute asynchronously}	
		\State ${y}_{j} \leftarrow f_{j}(\mathbf{x}^*)$ on $p_m$ where $f_{j}(\mathbf{x})$ is accuracy for $j$th epoch 
		\State {$\mathcal{A}\leftarrow\mathcal{A}\cup(j,{y}_{j})$}
		\State ${y}^{*} \leftarrow \max({y}^{*},{y}_{j})$
		\State {$\mathcal{H}\leftarrow\mathcal{H}\cup(\mathbf{x}^*, g({y}^{*}, \alpha), \mathcal{A})$} \ \ \ \ \ \ \ \ \ \ \ \ \ \ \ \ \ \# \textit{in-progress history update after loss transformation}
		\If{$j \in l(M, \beta)$}  
		\If {${y}^{*} < h(j, \mathcal{A},\mathcal{H}, \beta)$} 
		\State \textbf{break} \ \ \ \ \ \ \ \ \ \ \ \ \ \ \ \ \ \ \ \ \ \ \ \ \ \ \ \ \ \ \ \ \ \ \ \ \ \ \ \ \ \ \ \ \ \ \ \ \ \ \ \ \ \ \ \ \ \ \ \ \ \ \ \ \ \ \ \ \ \ \ \ \ \ \ \ \ \ \ \ \ \ \ \ \ \ \ \ \# \textit{early termination}
		\EndIf			
		
		\EndIf 
		\EndFor
		\If{${y}^{*} > c$}
		\State \textbf{break} \ \ \ \ \ \ \ \ \ \ \ \ \ \ \ \ \ \ \ \ \ \ \ \ \ \ \ \ \ \ \ \ \ \ \ \ \ \ \ \ \ \ \ \ \ \ \ \ \ \ \ \ \ \ \ \ \ \ \ \ \ \ \ \ \ \ \# \textit{finish when desired accuracy achieved}
		\EndIf		
		\EndFor
	\end{algorithmic}
\end{algorithm}

\subsection*{Hybrid log transformation}

\begin{align}
g(y, \alpha)= 
\begin{cases}
y              & \text{if } y < 1 - \alpha \\  
1 - \operatorname{log}(1 - y)  + (\operatorname{log}(\alpha) - \alpha),& \text{otherwise}
\end{cases}
\end{align}
Note that ($\operatorname{log}(\alpha)$ - $\alpha$) is added to make $g(y)$ a continuous function. In our work, we use $\alpha=0.3$. When $f(\mathbf{x})$ is not an accuracy, its range can be scaled to come within [0,1].

\subsection*{Compound termination rule}

\begin{figure}[h]
	\begin{center}
		\centerline{\includegraphics[width=0.7\columnwidth]{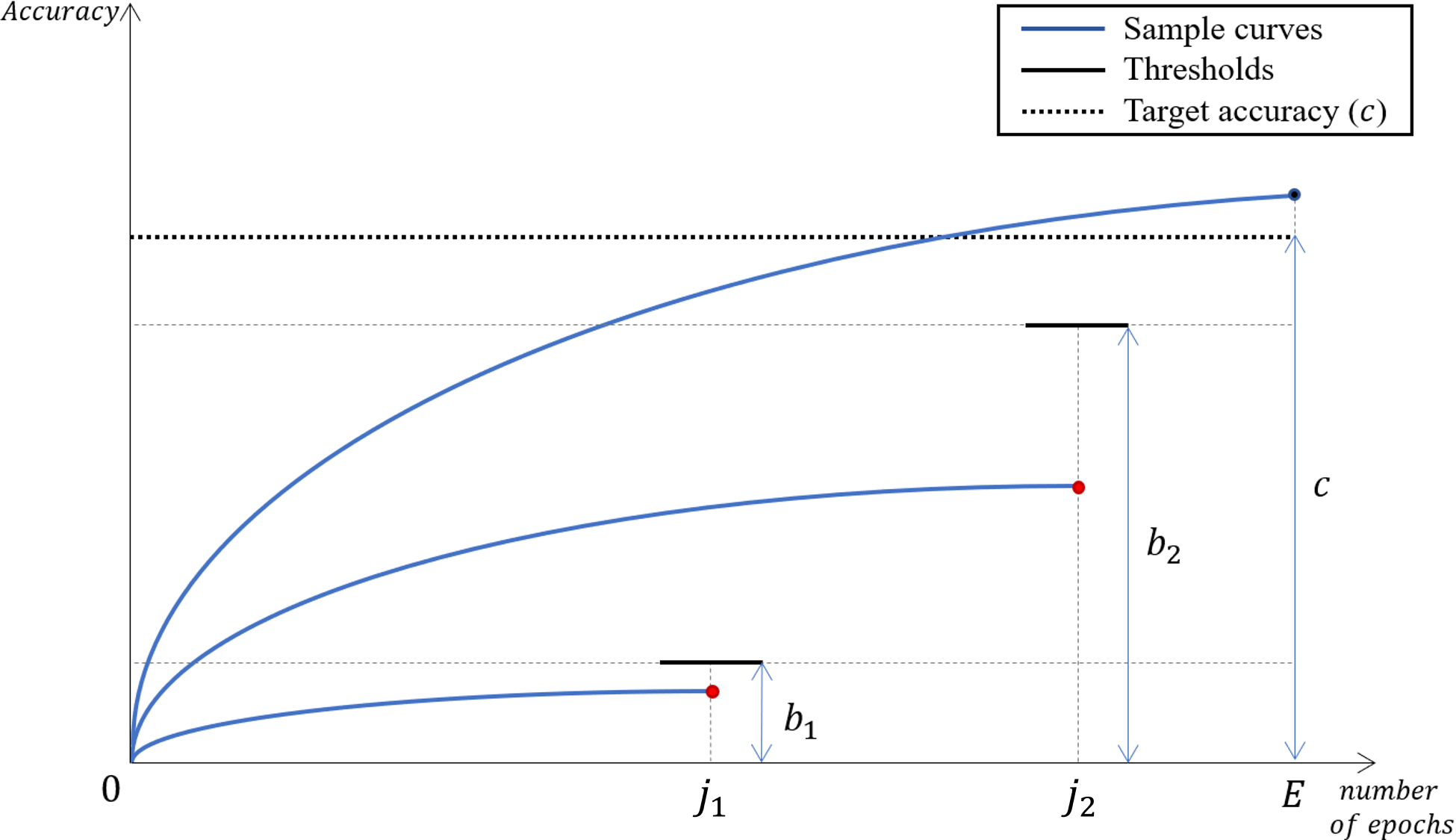}}
		
		\caption*{Figure A1. An illustration of our Compound Rule with 3 sample curves. Epoch $j_1$ and $j_2$ are checkpoints that determine whether a given $\mathbf{x}$ should be terminated, and $c$ is target accuracy. Here, they are $j_1=\lfloor 0.5E\rfloor,\;j_2=\lfloor (1-\beta)E\rfloor$. $\mathbf{x}$ is terminated at epoch $j_1$ if its performance at epoch $j_1$ is below $b_1$ which is computed referring to all $\mathbf{x}\in{\mathcal{H}}$. The second checkpoint, epoch $j_2$, applies to those that survived the first checkpoint for which the same logic applies but with $b_2$ whose computation refers only the history of those that survived the checkpoint-1.}
	\end{center}
\end{figure}

We designed an early termination rule and named as the Compound Rule (CR). 
We first introduce two formulaic tools required for computing function $l(E, \beta)$ and $h(j, \mathcal{A}, \mathcal{H}, \beta)$ which constitute our CR. Function $F_X(x)$ refers to the cumulative distribution function (CDF) for the distribution of a given random variable $X$. The rule's design constrains the range of $\beta$ to $\beta\in{(0.0,\;0.5]}$.
\begin{align}
	F_X(x) &=P(X\leq{x}) \label{alg(2)} \\ 
	\bar{f}_{i:j}({\mathbf{x}}) &=\frac{1}{j}\sum^j_{k=i}f_{k}({\mathbf{x}})\;\;\;\text{if}\;i\leq{j}\
\end{align}
Next, we show $l$ which returns the two checkpoints' epoch locations (epoch $j_1$ and $j_2$) and define $S$, that holds the history for configurations that survived the checkpoint $j_1$ computed by function $l$.  
\begin{align}
	l(E, \beta) &=\{\lfloor0.5E\rfloor,\;\lfloor(1-\beta)E\rfloor\} = \{j_1,\;j_2\} \\
	S &=\{s\in{\mathcal{H}}\mid\text{for}\;\mathcal{A}\in{s},\; n(\mathcal{A})>j_1\}
\end{align}
Using the above, we show $h$ which returns one of two thresholds. If $j\leq{j_1}$, $h$ returns the threshold for epoch $j_1$ (checkpoint-1), and otherwise returns the threshold for epoch $j_2$ (checkpoint-2).  
\begin{align}
	h(j, \mathcal{A}, \mathcal{H}, \beta)=
	\begin{cases}
		F^{-1}_{X}(\beta)            & \text{if } j \leq j_1 \\  
		F^{-1}_{X}(1-\beta)	         & \text{otherwise}
	\end{cases}
	\;\;\;\text{for,} \\
	X=
	\begin{cases}
		\{x\mid x=\bar{f}_{1:j_1} ({\mathbf{x}})\;\text{for}\;\mathbf{x}\in{\mathcal{H}}\} & \text{if } j \leq j_1 \\  
		\{x\mid x=\bar{f}_{j_1:j_2} ({\mathbf{x}})\;\text{for}\;\mathbf{x}\in{\mathcal{S}}\}	   & \text{otherwise}
	\end{cases}
\end{align}

Under a hypothetical environment where mapping early terminated configurations with intermediate performances does not penalize BO's modeling and per-epoch training time is fixed at $\frac{1}{M}$, larger $\beta$ is  observed to increase the number of configurations a BO process expectedly encounters per unit time. 
This is expressed as $q(\beta)$ below: 
\begin{gather}
q(\beta)=(0.5\beta+(1-\beta)^3+(1-\beta)\beta)^{-1}
\end{gather}
$q(\beta)$ initially increases almost linearly with $\beta$ but starts to wane away from this linear trajectory from about $\beta=0.2$. Though this conditioned observation suggests choosing $\beta\in{[0.05,\;0.20]}$, such condition would not hold for practical BO processes, making deterministic assumptions difficult. $\beta$ may be tuned based on practical user constraints.

%% file: supple-b.tex
\section*{Appendix B. \quad Early Termination Rule (ETR)}

\subsection*{Structural components for ETR}

ETRs rely on parameters that control several smaller knobs composing such structures. 
Traditionally, works in ETRs explore the theoretical and empirical robustness of the ETR as a single compound but not these knobs that comprise that compound. Comprehension of these knobs and their inner-workings is likely to further our understanding of early termination and potentially provide intuitions for ETR designs. In this part of our study, we attempt to observe and objectify some common knobs that many ETRs may share. 

We can annotate that an ETR evaluates works on a learning curve model $L_{\Theta}$ where $\Theta$ defines $L$ with preset features $\{\theta_1, \theta_2 ...\theta_k\}$ for $k\in\mathbb{N}$. 
Each $\theta_k$ refers to the $L$'s features which include, but are not excluded to, the maximum budget for training a configuration, whether $L$ is a regression model, refers to configuration history for adaptation, or uses extrapolated fantasy values for modeling. 
$L_{\Theta}$ takes 2 vector inputs: the selected hyperparameter configuration $\mathbf{x}\in\mathbf{X}$ and a design parameter vector $v\in\mathbf{V}$ to decide whether to terminate $\mathbf{x}$. 
If $\mathbf{T}=\{0,1\}$ where 0 means termination and 1 means no termination,

\begin{gather}
L_{\Theta}:\mathbf{X}\times \mathbf{V} \rightarrow \mathbf{T}
\end{gather}
whose domain is a Cartesian product. $\mathbf{X}$ would be forced to be an ordered vector space if some $\theta_k\in\Theta$ dictates $L_{\Theta}$ to refer to the configuration history. 

Number of elements in $v$ vary according to the proposed ETRs but are common in that they control several inner knobs at once. We observe and objectify 4 inner knobs and with them build on this equation.

\begin{itemize}

	\item\textbf{Observation Start Point $s$:} This refers to a point from which the learning curve is meaningfully observed to compute the threshold. $s$ is usually defined as a position relative to the maximum trainable epochs or resources allocated to $\mathbf{x}$. Its value is observed to be proportional to the maximum epochs and is usually expressed as its percentile value. 
	
	\item\textbf{Observation End Point $e$:} This refers to a point until which the learning curve is meaningfully observed to compute the threshold. $e$ is also observed to be defined proportional to the maximum epochs and together with point $s$ determines the scope of learning curve information that the ETR employs for evaluating $\mathbf{x}$. $s$ by definition cannot exceed $e$. Cases where $s=e$ is when the ETR uses the performance at that epoch alone to evaluate $\mathbf{x}$. 
	
	\item\textbf{Observation Checkpoint $j$:} This refers to a point where the learning curve is ultimately evaluated for termination. $j$ is also observed to be defined proportional to the maximum epochs. $e$ by definition cannot exceed $j$. Many cases set $j=e$ so that the evaluation of $\mathbf{x}$ is made right after the end of observation. For our study, we fix $j$ to $e$.
	
	\item\textbf{Termination Threshold $h$:} This refers to a knob that controls the threshold aggressiveness at the point $e$ with which the configuration is terminated or not. $h\in[0,1]$ where $h=0$ essentially equals to no ETR applied and $h=1$ indiscriminately terminates all configurations at point $e$. Given $v$, $h$ with a vector input of performances defined by $s$ and $e$ determines how the $L_{\Theta}$ should evaluate $\mathbf{x}$ at point $j$. 
	
	
\end{itemize}

An ETR that has a single checkpoint to evaluate $\mathbf{x}$ requires one value for each knob. An ETR that has $n$ checkpoints would require ordered vectors for each knob at length $n$ so that checkpoint-$i(i\leq n)$ corresponds to $s_i$, $e_i$, $j_i$, and $h_i$.

\subsection*{Survivor Rank Regret}

\textit{Survivor rank regret} is calculated as the mean value of \textit{rank regrets} for configurations that are not terminated by the ETR. We first explain \textit{rank regrets} and then \textit{survivor rank regret}.

\textit{Rank regret} is an adaption of the \textit{`immediate regret'}~\citep{shah2015parallel} whose original definition is $r_t=|f(\tilde{\mathbf{x}}_t)-f(\mathbf{x}^*)|$, 
where $\mathbf{x}^*$ is the known optimizer of $f$ and $\tilde{\mathbf{x}}_{t}$ is the recommendation of an algorithm after $t$ batch evaluations. \textit{Rank regret} uses ranks as the basic unit of measurement instead of $f(\mathbf{x})$ where the ranks are derived by using the pre-evaluated configuration tables generated with the Sobol sequences.
\textit{Rank regret} is used for our experiment instead because our work attempts to compare between datasets to check for the algorithm's robustness. `Immediate regret' is not suitable for such inter-dataset analysis because each dataset has different characteristics and different distributions of achievable performance and therefore the scale of $r_t$ values about BO vary greatly about datasets. Amount of change observed in $r_t$ for before and after applying some ETR to BO is thereby useless for inter-dataset comparison and analysis.
    
\textit{Rank regret}, and thereby \textit{survivor rank regret}, on the other hand is able to deliver information on relative performance improvement. \textit{Rank regret} first lines up the configurations according to their best performances and ranks the best among those configurations as the first. If we define the rank of $\mathbf{x}$ in the table as $rank(\mathbf{x})$, we normalize its difference with $rank(\mathbf{x}^*)$ which by definition is 1 so we get \textit{Rank regret} $=\frac{rank(\mathbf{x})-rank(\mathbf{x}^*)}{n(data\;table)}$ where $n(data\;table)$ refer to the number of configurations in the table. Our pre-evaluated configurations for each observed dataset contained 20,000 configurations, so $n(data\;table)$ for this work would simply equal 20,000.

\textit{Survivor rank regret} here is simply the mean value of \textit{rank regret} values for a subset of observed configurations that were not terminated at any checkpoint of the ETR and was fully trained to the maximum allocated budget for the configuration. This metric is intended to evaluate whether the configurations that the ETR decided to be worth fully training were, on average, better than those chosen without the ETR. Smaller \textit{survivor rank regret} values would indicate that the surviving pool of configurations are those with high performances and therefore more likely reach some given goal accuracy. If they are not as good as the baseline average \textit{rank regret} of an BO with no ETR, then that ETR would be of no help, or even harming, the BO. As this metric is intended for inter-dataset comparison, small inter-dataset variance values for respective \textit{survivor rank regret} values would signify that the ETR of concern is more likely robust in regards of ETR's contribution to BO.

\subsection*{A Delayed Observation Start Point for CR's Checkpoint-2}

\begin{table}[!b]
	\centering
	
	\caption*{Table B1. Each cell shows $\mu\pm\sigma$ of the respective median \textit{survivor rank regret} values for 3 pre-evaluated surrogate configurations(MNIST-LeNet2, PTB-LSTM, and CIFAR10-VGG benchmark) for 50 trials with our ETR on the cell's setting(row:$s_2$, column:$e_2$). As DEEP-BO uses $\beta=0.1$, this table sets the threshold at the 90th percentile value for comparison.}
	
	\label{tab_survrankreg}
	\resizebox{0.9\columnwidth}{!}{
		\begin{threeparttable}
			\begin{tabular}{cccc|ccc}
				\toprule
				
				\textbf{BO}   & \multicolumn{3}{c|}{\textbf{GP-EI}}              & \multicolumn{3}{c}{\textbf{RF-EI}}                     \\ \cmidrule(l){2-7} 
				&  \textbf{50\%}   & \textbf{70\%}   & \textbf{90\%}   & \textbf{50\%}   & \textbf{70\%}   &      \textbf{90\%}   \\ \cmidrule(l){1-7}
				\textbf{0\%}  & $13.2\pm5.1$\% & $13.2\pm5.2$\% & $12.7\pm4.2$\% & $4.8\pm4.8$\% & $4.2\pm4.0$\% &       $3.7\pm2.9$\% \\
				\textbf{10\%} & $11.3\pm3.2$\% & $11.2\pm3.4$\% & $10.6\pm2.5$\% & $2.9\pm2.6$\% & $3.0\pm2.3$\% &       $2.9\pm2.2$\% \\
				\textbf{20\%} & $9.4\pm1.5$\% & $9.6\pm1.2$\% & $9.2\pm1.1$\% & $1.5\pm0.9$\% & $1.9\pm1.2$\% &       $1.7\pm1.0$\% \\
				\textbf{30\%} & $8.0\pm1.1$\% & $8.8\pm0.6$\% & $8.6\pm0.8$\% & $1.1\pm0.6$\% & $1.2\pm0.7$\% &       $1.3\pm0.7$\% \\
				\textbf{40\%} & $7.1\pm0.3$\% & $7.4\pm0.5$\% & $7.9\pm0.5$\% & $0.5\pm0.4$\% & $0.8\pm0.3$\% &       $0.9\pm0.4$\% \\ \bottomrule
				
			\end{tabular}
		\end{threeparttable}
	}
\end{table}

This part of the work incorporates the structural framework introduced in this appendix to observe experiment results summarized in Table B1 and suggest the motivations for choosing a delayed observation start point ($s_2=j_1=\lfloor{E/2}\rfloor$) for checkpoint-2. We note that the results in this experiment is a full version of CR but closer to MSR particularly intended for CR's checkpoint-2. This experiment limits the focus to observing the effects of the change in $s_2$ on the \textit{survivor rank regret} which acted as one motivation for the design choice of CR's checkpoint-2.

As shown in Table B1, $s_2\in\{0.0, 0.1, 0.2, 0.3, 0.4\}$ and $e_2\in\{0.5, 0.7, 0.9\}$ (those corresponding to checkpoint-1 is $s_1$ and $e_1$) together form the scope of epochs observed in producing the 90th percentile threshold at $e_2=j_2$ for checkpoint-2 (subscript 2 indicates that $e$ and $j$ are those corresponding to checkpoint-2). Among these 2 knobs, we wish to focus on $s_2$ in particular where for the given column, the inter-dataset $\mu$ and $\sigma$ values for \textit{survivor rank regret} strictly decrease as $s_2$ increases. This work considers this dynamic to be an important marker for using a delayed $s_2$ for checkpoint-2 in regards of enhancing both performance and robustness. Observation of decreasing $\mu$ indicates that this algorithm (with a delayed $s_2$) improves BO's potential performance for datasets overall and the observation with decreasing $\sigma$ indicates greater robustness among datasets. Though the dynamic between $s_2$ and $\mu$ does contribute to the reasoning for our choice of $s_2$ for checkpoint-2, our work puts more emphasis on the dynamic between $s_2$ and $\sigma$.

Smaller $\sigma$ among datasets is an indicator that the effects of an algorithm on BO, whether good or bad (here is the former considering the observations with $\mu$ values), is stable and therefore more likely to act in such steady manner even for unseen datasets. This observation with decreasing $\sigma$ can also be intuitively inferred, as exemplified in Figure 4, considering the fact that the overall maximum deviations in performance are smaller (though that amount varies according to the type of curves) for latter epochs. As stated in the paper, occasional disasterous failures are intolerable and thereby the quality of some algorithm to perform well and steadily is crucial for practical DNN HPO. Therefore, the early termination rule for our DEEP-BO algorithm chooses to have a more delayed observation start point. As checkpoint-2 considers only the configurations that survived checkpoint-1 at epoch $j_1$, the observation start point for checkpoint-2 here is set at $j_1$ for convenience. 


\subsubsection*{ETR performance comparison}

Evaluating the ETR performance only, we compare our ETR, which here we name as Compound rule(CR), to existing ETRs such as Median Stopping Rule(MSR)~\citep{golovin2017google}, Learning Curve Extrapolation(LCE)~\citep{domhan2015speeding}. 
We select GP-EI as a default BO algorithm which will be accelerated.
We evaluated BOHB instead of HB. BOHB is an extended version of HB, which works like an ETR by dynamic resource allocation manner, and also this algorithm already utilizes the other BO. 
\cite{baker2017accelerating} is excluded in this study because this requires a long warm-up time to train the regression model, which is impractical to our benchmark problem.     

\begin{figure}[ht]
	\centering
	\begin{subfigure}[t]{.48\textwidth}
		\includegraphics[width=\columnwidth]{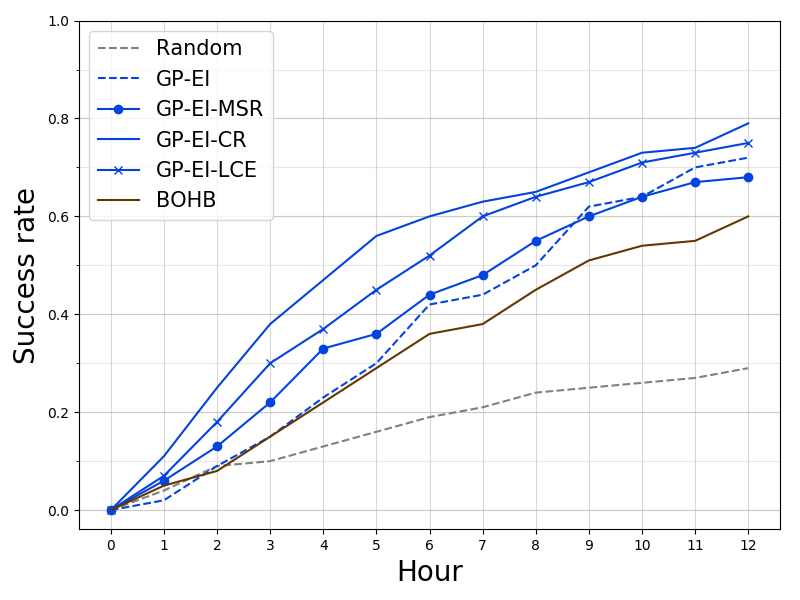}
		\caption{MNIST-LeNet1}
		\label{fig6:a}
	\end{subfigure}%
	\begin{subfigure}[t]{.48\textwidth}
		\includegraphics[width=\columnwidth]{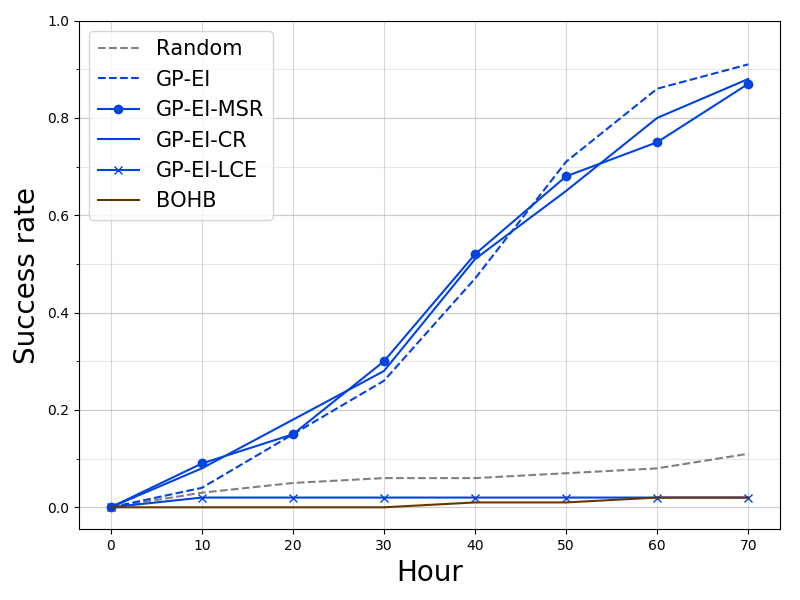}
		\caption{CIFAR10-ResNet}
		\label{fig6:b}
	\end{subfigure}%
	\caption*{Figure B1. \textit{Success rate} of ETRs achieving Top-10 accuracy. ETRs are used on GP-EI. MSR stands for Median Stopping Rule by \cite{golovin2017google}. CR stands for our compound termination rule. LCE stands for Learning Curve Extrapolation by \cite{domhan2015speeding}.}
\end{figure}

As shown in Figure B1, our CR performs better in MNIST-LeNet1 or similar in CIFAR10-ResNet than MSR. 
However, LCE and BOHB which performs similarly with CR in MNIST-LeNet1 perform worst in CIFAR10-ResNet. 
Especially, the ETRs which performs worse than random means that they failed severely.
We tried best to optimize their termination criteria but, we could not make them robust in this benchmark. 
These model-based algorithms assume that the asymptotic performance of a given configuration observing with just partial points in a learning curve can be highly predictable. This may cause their prediction failures and also modeling failure of BO. 
However, model-free algorithms such as MSR seem to perform robustly in any benchmark.

%% file: supple-c.tex
\section*{Appendix C.\quad Experiment Design}

\subsection*{Benchmark algorithms}
\label{subsec:algorithms}
For the modeling algorithms, we adopted two of the most popular modeling algorithms of BO - GP and RF. Specifically, we have closely followed the implementations of Spearmint for GP and SMAC for RF. For GP, we used the automatic relevance determination(ARD) Mat\'ern 5/2 kernel and a Monte Carlo estimation of the integrated acquisition as suggested in \cite{snoek2012practical}. 
For RF, we used the hyperparameter setting where the number of trees and the number of minimum items in a split are set to 50, 2, respectively.
We also one-hot encoded for categorical hyperparameter values to avoid unexpected distortion of modeling.

For the acquisition function, we utilized the three basic approaches - EI, PI, and UCB. By considering all possible combinations between the modeling algorithms and the acquisition functions, we ended up with six different algorithms. For DEEP-BO, we have adopted the six combinations as individual algorithms and implemented the diversification strategies. 
The algorithms that are evaluated are summarized below. 
\begin{itemize}
	\item The random algorithm~\citep{bergstra2012random} 
	\item Six individual BO algorithms. Cartesian product by two modeling algorithms~\citep{snoek2012practical, hutter2011sequential} and three acquisition functions(Probability of Improvement, Expected Improvement, and Upper Confidence Bound)~\citep{kushner1964new, mockus1978toward,srinivas2009gaussian}: GP-EI, GP-PI, GP-UCB, RF-EI, RF-PI, and RF-UCB
	\item GP-Hedge introduced by~\cite{hoffman2011portfolio}, where GP-EI, GP-PI, GP-UCB are used as the three arms of multi-armed bandit
	\item RF-EI using Learning Curve Extrapolation (LCE)~\citep{domhan2015speeding}.
	\item GP-EI using Median Stopping Rule (MSR)~\citep{golovin2017google}
	\item Hyperband (HB)~\cite{li2017hyperband}
	\item BOHB which allocates resources as an infinite bandit-based approach but, it samples better than random using TPE~\citep{falkner2018bohb}.
	\item Our method DEEP-BO.
\end{itemize}
Random algorithm was included because it is frequently used to replace grid-search, but typically its performance is much worse than the others for DNN tasks. This will become obvious in the result plots.

\subsection*{Hyperparameter configuration space}

\begin{table}[H]
	\centering
	\caption*{Table C1: The hyperparameter spaces of LeNet on MNIST dataset.}
	\label{tab-d1}
	
	\resizebox{\textwidth}{!}{	
		\begin{tabular}{lp{5.7cm}cccc}
			\toprule
			\multirow{2}{*}{\textbf{Architecture}} & \multirow{2}{*}{\textbf{Name}}             & \multirow{2}{*}{\textbf{Type}} & \multicolumn{2}{c}{\textbf{Range}}                                                                                \\
			\cmidrule(l){4-5}
			
			&                                            &                                & LeNet1              & LeNet2                               \\
			\midrule
			\multirow{12}{*}{LeNet-5}              & \pbox{20cm}{number of first convolution\\ kernels}        & discrete                            & {[}1, 350{]}                     & {[}1, 350{]}                                   \\
			
			\cmidrule(l){2-5}
			& \pbox{10cm}{first pooling size}                         & discrete                            & {[}2, 3{]}                    &    -                                            \\
			
			\cmidrule(l){2-5}
			& \pbox{10cm}{number of second convolution\\ kernels}       & discrete                            & {[}1, 350{]}                  & {[}1, 350{]}                                   \\
			
			\cmidrule(l){2-5}
			& \pbox{10cm}{second pooling size}                        & discrete                            & {[}2, 3{]}                    &    -                                            \\
			
			\cmidrule(l){2-5}
			& \pbox{10cm}{number of neurons in fully\\ connected layer} & discrete                            & {[}1, 1024{]}                 & {[}1, 1024{]}                                  \\
			
			\cmidrule(l){2-5}
			& \pbox{10cm}{square length of the convolution\\ kernel}       & discrete                            & {[}2, 10{]}                   &   -                                             \\
			
			\cmidrule(l){2-5}
			& learning rate                              & continuous                          & \pbox{10cm}{{[}0.0001, 0.4{]}\\(log scale)} & \pbox{10cm}{{[}0.0001, 0.4{]}\\(log scale)}                  \\
			
			\cmidrule(l){2-5}
			& L2 regularization factor                   & continuous                          & {[}0.0, 1.0{]}                & {[}0.0, 1.0{]}                                               \\
			
			\cmidrule(l){2-5}
			& drop out rate                              & continuous                          & {[}0.0, 0.9{]}                &  {[}0.0, 1.0{]}                                              \\
			
			\cmidrule(l){2-5}
			& activation function type                   & categorical                            &     -                          & \pbox{10cm}{ReLU, tanh,sigmoid, \\eLU, leaky ReLU}                \\
			
			\cmidrule(l){2-5}
			& optimizer type                             & categorical                            &      -                         & \pbox{10cm}{AdaDelta, AdaGrad, Adam, \\GD, Momentum, RMSProp} \\
			
			\cmidrule(l){2-5}
			& batch normalization                        & categorical                            &      -                         & \pbox{10cm}{Enable, Disable}                         \\
			
			\bottomrule
		\end{tabular}
	}
\end{table}

\begin{table}[H]
	\centering
	\caption*{Table C2: The hyperparameter spaces of VGGNet and ResNet on CIFAR10.}
	\label{tab-d3}
	\resizebox{\textwidth}{!}{
		
		\begin{tabular}{lp{8cm}cc}
			\toprule
			\textbf{Architecture} & \textbf{Name}             & \textbf{Type} & \textbf{Range}                                                                                \\
			
			\midrule
			\multirow{8}{*}{VGG}              & \pbox{10cm}{number of first convolution kernels}        & discrete                            & {[}8, 32{]}                      \\
			
			\cmidrule(l){2-4}
			
			& \pbox{10cm}{number of second convolution kernels}       & discrete                            & {[}32, 64{]}                     \\
			
			\cmidrule(l){2-4}
			& \pbox{10cm}{number of third convolution kernels}       & discrete                            & {[}64, 128{]}                     \\
			
			\cmidrule(l){2-4}
			& \pbox{10cm}{number of forth convolution kernels}       & discrete                            & {[}64, 128{]}                     \\
			
			\cmidrule(l){2-4}				
			& \pbox{10cm}{number of neurons in fully connected layer} & discrete                            & {[}10, 1000{]}                    \\
			
			\cmidrule(l){2-4}
			& \pbox{10cm}{square length of convolution kernel}       & discrete                            & {[}2, 3{]}                      \\
			\cmidrule(l){2-4}
			& learning rate                              & continuous                          & \pbox{10cm}{{[}0.0001, 0.4{]} \\(log scale)}  \\
			
			\cmidrule(l){2-4}
			& L2 regularization factor                      & continuous                          & {[}0.0, 1.0{]}                   \\
			
			\cmidrule(l){2-4}
			& activation function type                   & categorical                            & \pbox{10cm}{ReLU, tanh, eLU}                \\
			
			\cmidrule(l){2-4}
			& regularization method                        & categorical                            & \pbox{10cm}{None, Dropout, BatchNorm}        \\

			\midrule
			\multirow{7}{*}{ResNet}              & \pbox{10cm}{number of layers}       & discrete                            & 38, 44, 50, 56, 62, 70, 78\\
			
			\cmidrule(l){2-4}
			& \pbox{10cm}{training data batch size}       & discrete                            & 45, 90, 180, 360, 450\\
			\cmidrule(l){2-4}
			& \pbox{10cm}{data augmentation}       & categorical                           & True, False\\
			\cmidrule(l){2-4}
			& learning rate                              & continuous                          & \pbox{10cm}{{[}0.0001, 0.1{]} \\(log scale)}  \\
			
			\cmidrule(l){2-4}
			& \pbox{10cm}{momentum for MomentumOptimizer}       & continuous                            & {[}0.1, 0.9{]} \\
			
			\cmidrule(l){2-4}
			& \pbox{10cm}{weight decay }       & continuous                            & \pbox{10cm}{{[}0.00001, 0.001{]}\\(log scale)} \\
			\cmidrule(l){2-4}
			& \pbox{10cm}{batch normalization decay }       & continuous                            & \pbox{10cm}{{[}0.9, 0.999{]}} \\

			\bottomrule
		\end{tabular}
	}
	
\end{table}

\begin{table}[H]
	\centering
	\caption*{Table C3: The hyperparameter space of LSTM on PTB dataset. }
	\label{tab-d2}
	\resizebox{\textwidth}{!}{
		
		\begin{tabular}{lp{8cm}cc}
			\toprule
			\textbf{Architecture} & \textbf{Name}             & \textbf{Type} & \textbf{Range}                                                                                \\
			\midrule
			\multirow{12}{*}{LSTM}              & \pbox{10cm}{number of neurons in hidden layer}        & discrete                            & {[}10, 200{]}\\
			
			\cmidrule(l){2-4}
			
			& \pbox{10cm}{number of hidden layers}       & discrete                            & {[}1, 2{]}\\
			
			\cmidrule(l){2-4}
			& \pbox{10cm}{number of training steps}       & discrete                            & {[}10, 20{]}\\
			
			\cmidrule(l){2-4}
			& \pbox{10cm}{initial value of uniform random scale}       & continuous                            & {[}0.01, 0.1{]}\\
			
			\cmidrule(l){2-4}				
			& drop out rate                              & continuous                          & {[}0.0, 0.9{]} \\
			
			\cmidrule(l){2-4}
			& learning rate                              & continuous                          & \pbox{10cm}{{[}0.1, 1.0{]}} \\
			
			\cmidrule(l){2-4}
			& learning rate decay                   & continuous                          & {[}0.5, 1.0{]} \\
			
			\cmidrule(l){2-4}
			& \pbox{10cm}{gradient clipping by global normalization}                   & continuous                            & {[}5.0, 10.0{]} \\
			
			\cmidrule(l){2-4}
			& RNN training module                        & categorical                            & \pbox{10cm}{Basic, Block, cuDNN}\\
			\bottomrule
			
		\end{tabular}
	}
	
\end{table}

\begin{table}[H]
	\centering
	\caption*{Table C4: The hyperparameter space of VGGNet on CIFAR100 dataset.}
	\label{tab-d4}
	\resizebox{\textwidth}{!}{
		
		\begin{tabular}{lp{8cm}cc}
			\toprule
			\textbf{Architecture} & \textbf{Name}             & \textbf{Type} & \textbf{Range}                                                                                \\
			
			\midrule
			\multirow{10}{*}{VGG}              & \pbox{10cm}{number of first convolution kernels}        & discrete                             & {[}8, 32{]}                  \\
			
			\cmidrule(l){2-4}
			
			& \pbox{10cm}{number of second convolution kernels}       & discrete                           & {[}32, 64{]}\\
			
			\cmidrule(l){2-4}
			& \pbox{10cm}{number of third convolution kernels}       & discrete                           & {[}64, 128{]}\\
			
			\cmidrule(l){2-4}
			& \pbox{10cm}{number of forth convolution kernels}       & discrete                            & {[}64, 128{]}\\
			
			\cmidrule(l){2-4}				
			& \pbox{10cm}{number of neurons in fully connected layer} & discrete                            & {[}10, 1000{]}\\
			
			\cmidrule(l){2-4}
			& \pbox{10cm}{square length of convolution kernel}       & discrete                            & {[}2, 3{]} \\
			
			\cmidrule(l){2-4}
			& learning rate                              & continuous                          & \pbox{10cm}{{[}0.0001, 0.4{]}\\(log scale)}                  \\
			
			\cmidrule(l){2-4}
			& L2 regularization factor                     & continuous                          & {[}0.0, 1.0{]}                \\
			
			\cmidrule(l){2-4}
			& activation function type                   & categorical                            & \pbox{10cm}{ReLU, tanh,sigmoid, eLU} \\
			
			\cmidrule(l){2-4}
			& regularization method                        & categorical                            & \pbox{10cm}{None, Dropout, BatchNorm}                 \\
			
			\bottomrule
		\end{tabular}
	}
	
\end{table}


\subsection*{Surrogate configurations}

The performance of an HPO algorithm for a specific task is not deterministic. For example, GP-EI would output a different result for each trial of hyperparameter optimization. 
Therefore, repeated experiments are necessary to estimate the \textit{`success rate'} and \textit{`expected time'} to success in a credible way. 
However, each trial of a single HPO run requires a considerable number of DNN trainings (typically a few hundreds of DNN trainings), and thus it would be extremely time-consuming to repeat the experiments over multiple algorithms. 
In this regard, \citet*{eggensperger2015efficient} suggest that surrogate benchmarks are cheaper to evaluate and closely follow real-world benchmarks.

\begin{table}[!hb]
	\centering
	\caption*{Table C5: Statistics of pre-evaluated configurations.}
	\label{tab-a5}
	\resizebox{\textwidth}{!}{
		
		\begin{tabular}{p{7.5cm}|rrrrrrr}
			\toprule
			\textbf{Dataset}                                                                         & \multicolumn{2}{c}{MNIST}  & PTB &  \multicolumn{2}{c}{CIFAR10} & CIFAR100 \\
			\cmidrule(l){1-7}
			\textbf{Architecture}                                                                          & LeNet1 & LeNet2 & LSTM    & VGG  & ResNet  & VGG \\		
			\midrule
			Total evaluation time (day)                                                        & 88.1    & 55.5    & 87.3   & 75.2     & 385    & 82.1      \\
			Mean evaluation time (min)                                                         & 6.3     & 4       & 6.3    & 5.4      & 79     & 5.9       \\
			Global optimum (top accuracy)                                                 & 0.9939  & 0.9941  & 0.9008 & 0.8052   & 0.9368  & 0.5067    \\
			Target performance (top-10 accuracy)  & 0.9933  & 0.9922  & 0.8993 & 0.7836  & 0.9333  & 0.4596 \\
			\bottomrule   
		\end{tabular}
	}
\end{table}

To utilize surrogate benchmarks, we pre-selected a representative set of model configurations, pre-evaluated all the selected configurations, and saved their DNN training results (including the evaluated test performance and the consumed time) into a database. This process takes some time to complete because we need to complete DNN trainings of all the configurations. In our case, we had 20,000 configurations for five benchmarks and 7,000 for the other, and thus a total of 107,000 configurations worth of DNN training had to be completed for the six benchmarks. However, once completed, we were able to repeatedly perform HPO experiments using the database without requiring to resort to actual DNN evaluations every time they were necessary. With the pre-generated database, proper book-keeping can be performed to expedite the experiments. Using this methodology, we were able to evaluate each HPO algorithm 100 times for each benchmark task. 

In order to select \textit{`a representative set of configurations'}, we used Sobol sequencing method to choose the configurations. Sobol is a quasi-random low-discrepancy sequence in a unit hypercube, and its samples are known to be more evenly distributed in a given space than those from uniform random sampling. 
We set an appropriate range for each hyperparameter. Here, each range can be a real interval or a bounded set of discrete/categorical values. Then, if we were to say that we are concern with $d$ of hyperparameters, we cast this truncated hyperparameter space into a $d$-dimensional unit hypercube and generate a Sobol sequence within the hypercube. Finally, we train DNN for all configurations sampled by Sobol grids and save the necessary information such as respective training time and prediction accuracy. 
We collected these training results for six benchmarks on machines with NVIDIA 1080 TI GPU.

\subsection*{Hyperparameter configuration of BOHB}

Hyperparameter configurations for the MNIST-LeNet1 benchmark were applied with a minimum budget 1 and maximum budget 81. Other hyperparameters used for BOHB in our experiments are as follows; the scaling parameter $\eta$ is 3, the number of samples used to optimize the acquisition function is 64, the fraction of purely random configuration $\rho$ is 0.33, the minimum budget is 1, the maximum budget is 15. 

Hyperparameters, except the minimum and maximum budgets, followed the parameters used for the MNIST dataset in \cite{falkner2018bohb}. We were able to observe comparable results with those produced in the parameter settings provided by \cite{falkner2018bohb} when we experimented with the algorithm in our environmental settings by changing its hyperparameters for the given minimum and maximum budgets.

For the PTB-LSTM benchmark, we also used the hyperparameter configuration with minimum budget 1, maximum budget 15 and the scaling parameter $\eta$ as 3.   
For the CIFAR10-ResNet benchmark, the hyperparameters were applied with a maximum budget of 100 and the scaling parameter $\eta$ as 3. 
Other hyperpaprameters are the same as above.

%% file: supple-d.tex
\section*{Appendix D.\quad Benchmark Performance}

\subsection*{Using single processor}

Table B3 summarizes the performances of six individual algorithms and the diversification algorithms for the six benchmarks. 

Two types of performance metrics, \textit{success rate} and \textit{expected time}, were evaluated and those that performed best were shown in bold. For both metrics, the target performance $c$ was set as the lower bound of top-10 best performances found among the pre-evaluated results. 
All algorithms run until the best performance currently outperform the top-10.

As for the \textit{success rate} using a single processor, we list the \textit{success rate} after 12 hours for MNIST-LeNet, and PTB-LSTM, because most algorithms' average time for success is less than 12 hours. For a similar reason, the \textit{success rate} after 6 hours is considered for CIFAR10-VGG and CIFAR100-VGG. In the case of CIFAR10-ResNet, we list the \textit{success rate} after 90 hours.  
Note that the \textit{success rate} listed in Table D1 are cross-sectional snapshots of Figure 2 that show the time series of \textit{success rates}. 
On the mean performance of all benchmarks, the \textit{success rate} of DEEP-BO was 12\% superior to RF-PI, which was the best among the other algorithms.
DEEP-BO achieved 31\% of average improvements over the early-termination-disabled individual algorithms.
DEEP-BO is also 25\% better than GP-EI-MSR or GP-Hedge when compared to the other advanced algorithms.
Compared to random, the baseline algorithm, it's improved by 85\%.

Regarding \textit{expected time}, DEEP-BO achieved 60\% of time reductions over the early-termination-disabled individual algorithms.
DEEP-BO decreased 6.5 hours compared to the second fastest GP-EI-MSR.

\begin{table*}[ht]
	\centering
	\caption*{Table D1. Summary of single processor performance achieving top-10 accuracy.} 
	\label{tab-b1}
	\resizebox{\columnwidth}{!}{
	\begin{tabular}{l|p{3.5cm}|c|c|c|c|c|c}
		\toprule
		\textbf{Measure} & \textbf{Benckmark} & \textbf{GP-EI} & \textbf{GP-PI} & \textbf{GP-UCB} & \textbf{RF-EI} & \textbf{RF-PI} & \textbf{RF-UCB} \\ 
		\midrule
		\multirow{7}{*}{\begin{tabular}[c]{@{}l@{}}Success\\     rate\end{tabular}}
		& MNIST-LeNet1 \space\space\space(12h)     & 34\%           & 34\%           & 29\%            & 48\%           & \textit{61\%}           & 41\%             \\  
		& MNIST-LeNet2 \space\space\space(12h)    & 49\%           & 29\%            & 48\%             & 63\%           & 57\%           & \textit{65\%}              \\  
		& PTB-LSTM \space\space\space\space\space\space\space\space\space(12h)    & 89\%           & 83\%           & 86\%            & 68\%           & \textit{92\%}           & 89\%                 \\  
		& CIFAR10-VGG \space\space\space\space\space(6h)        & 50\%           & 43\%           & 46\%            & \textit{77\%}            & \textit{77\%}           & 75\%             \\  
		& CIFAR10-ResNet (90h)    & 91\%           & 86\%           & \textit{92\%}            & 69\%           & 89\%           & 86\%              \\  
		& CIFAR100-VGG \space\space\space(6h)   & 87\%           & 81\%           & 79\%            & 96\%           & \textit{98\%}           & 97\%             \\
		\cmidrule(l){2-8}
		& Mean   & 67\%           & 59\%           & 63\%            & 70\%           & \textit{79\%}           & 76\%        \\ 	 
		
		\midrule
		\multirow{7}{*}{\begin{tabular}[c]{@{}l@{}}Expected\\     time \\(hour)\end{tabular}} 
		
		& MNIST-LeNet1     & $53.1\pm74.4$           & $56.9\pm81.5$           & $67.5\pm84.4$            & $43.2\pm59.7$           & $27.8\pm65.2$           & $36\pm51.7$             \\  
		&  MNIST-LeNet2    & $29.6\pm46.7$           & $77.9\pm76.6$            & $25.3\pm27.5$             & $16.5\pm23.8$           &  $45.4\pm78.5$            & $20.4\pm34.3$            \\  
		& PTB-LSTM     & \space$8.2\pm3.4$           & \space$7.9\pm4.6$           & \space$8.6\pm3.4$            & $25.8\pm61.4$           & \space$6.4\pm3.4$           & \space$7.4\pm3.5$               \\  
		& CIFAR10-VGG         & $7.4\pm4.3$           & $7.2\pm3.5$           & $8.1\pm4.7$            & $4.7\pm3.2$            & $4.4\pm3.8$           & $4.4\pm3.5$                  \\  
		& CIFAR10-ResNet     & $44.1\pm28.9$           & $57.2\pm32.4$           & $47.5\pm24.6$            & $78.5\pm59.6$           & $53.5\pm35.2$           & $55.7\pm30.4$            \\  
		& CIFAR100-VGG  & $4.1\pm1.6$           & $4.1\pm2.1$           & $6.7\pm11.7$            & $3.5\pm6.3$           & $3.2\pm7.2$           & $3.1\pm5.5$          \\ 
		\cmidrule(l){2-8}
		& Mean  & $24.4$           & $35.2$           & $27.3$            & $28.7$           & $23.4$          & $21.2$     \\ 	
		\cmidrule(l){1-8}
		
	\end{tabular}		
	}
	\resizebox{\columnwidth}{!}{
	\begin{tabular}{l|p{3.5cm}|c|c|c|c|c|c}
		\textbf{Measure} & \textbf{Benckmark} & \textbf{Random} & \textbf{GP-Hedge} & \textbf{GP-EI-MSR} & \textbf{RF-EI-LCE} & \textbf{BOHB} & \textbf{DEEP-BO} \\ 
		\midrule
		\multirow{7}{*}{\begin{tabular}[c]{@{}l@{}}Success\\     rate\end{tabular}}
		& MNIST-LeNet1 \space\space\space(12h)     & 7\%           & 32\%           & 32\%            & 36\%           & 12\%           & \textbf{81\%}             \\  
		& MNIST-LeNet2 \space\space\space(12h)    & 12\%           & 39\%            & 48\%             & 54\%           & 1\%           & \textbf{82\%}              \\  
		& PTB-LSTM \space\space\space\space\space\space\space\space\space(12h)    & 3\%           & 85\%           & 89\%            & 41\%           & 29\%           & \textbf{100\%}                 \\  
		& CIFAR10-VGG \space\space\space\space\space(6h)        & 5\%           & 53\%           & 48\%            & \textit{78\%}            & 0\%           & \textbf{80\%}             \\  
		& CIFAR10-ResNet (90h)    & 7\%           & 91\%           & \textit{94\%}            & 0\%           & 4\%           & \textbf{100\%}              \\  
		& CIFAR100-VGG \space\space\space(6h)   & 2\%           & 92\%           & 85\%            & -\%           & 4\%           & \textbf{100\%}             \\
		\cmidrule(l){2-8}
		& Mean   & 6\%           & 66\%           & 66\%            & 42\%           & 8\%           & \textbf{91\%}        \\ 	 
		
		\midrule
		\multirow{7}{*}{\begin{tabular}[c]{@{}l@{}}Expected\\     time \\(hour)\end{tabular}} 
		
		& MNIST-LeNet1     & $169.1\pm142.6$           & $17.0\pm8.2$           & $16.3\pm8.6$            & $16.2\pm8.3$           & $20.7\pm6.4$           & \boldmath$8.2\pm8.4$             \\  
		&  MNIST-LeNet2    & $99.9\pm86.1$           & $29.3\pm31.2$            & $24.2\pm30.4$             & $41.4\pm66.6$           &  $35.8\pm12.1$            & \boldmath$7.6\pm10.4$            \\  
		& PTB-LSTM     & \space$188.3\pm169.2$           & \space$8.3\pm3.6$           & \space$7.9\pm2.8$            & $69.2\pm104.2$           & \space$28.3\pm22.7$           & \space\boldmath$4.8\pm2.2$               \\  
		& CIFAR10-VGG         & $138.2\pm137.6$           & $6.6\pm4.0$           & $6.8\pm4.3$            & $4.1\pm2.9$            & $29.4\pm1.8$           & \boldmath$3.4\pm2.5$                  \\  
		& CIFAR10-ResNet     & $756.1\pm724.8$           & $50.6\pm29.6$           & $43.8\pm25.4$            & $120.5\pm0.3$           & $129.8\pm17.8$           & \boldmath$38.4\pm17.3$            \\  
		& CIFAR100-VGG  & $195.8\pm155.0$           & $4.1\pm2.1$           & $4.2\pm3.0$            & $-\pm-$           & $24.8\pm9.9$           & \boldmath$1.9\pm0.9$          \\ 
		\cmidrule(l){2-8}
		& Mean  & $259.9$           & $25.7$           & $17.2$            & $50.3$           & $44.8$          & \boldmath$10.7$     \\ 	
		\cmidrule(l){1-8}
		
	\end{tabular}		
}
\end{table*}

\newpage

\begin{figure}[H]
	\centering
	\begin{subfigure}[b]{.45\textwidth}
		\includegraphics[width=\columnwidth]{figs/fig-mnist2-s2.png}
		\caption{MNIST-LeNet1}
		\label{figs1:a}
	\end{subfigure}
	~ 
	\begin{subfigure}[b]{.45\textwidth}
		\includegraphics[width=\columnwidth]{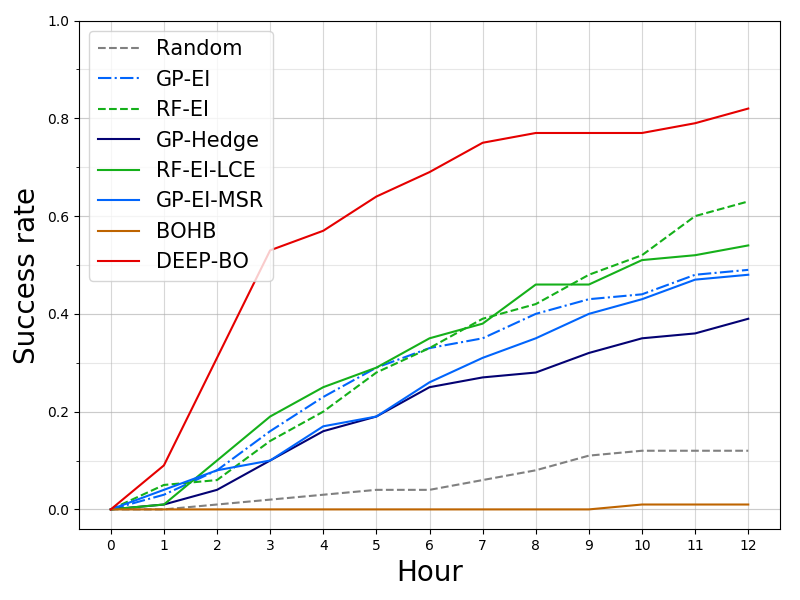}
		\caption{MNIST-LeNet2}
		\label{figs1:b}
	\end{subfigure}
	~ 
	\begin{subfigure}[b]{.45\textwidth}
		\includegraphics[width=\columnwidth]{figs/fig-ptb-s2.png}
		\caption{PTB}
		\label{figs1:c}
	\end{subfigure}
	~ 
	\begin{subfigure}[b]{.45\textwidth}
		\includegraphics[width=\columnwidth]{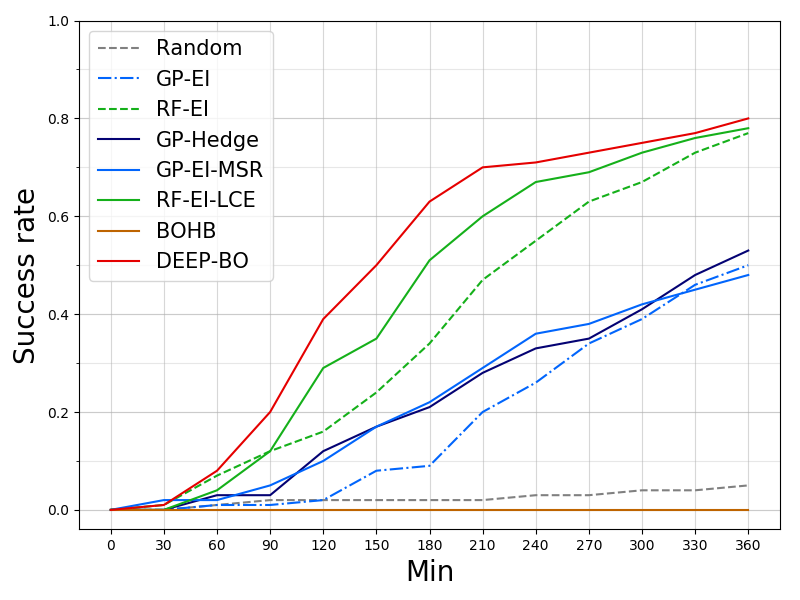}
		\caption{CIFAR10-VGG}
		\label{figs1:d}
	\end{subfigure}
	~ 
	\begin{subfigure}[b]{.45\textwidth}
		\includegraphics[width=\columnwidth]{figs/fig-cifar10-resnet-s2.png}
		\caption{CIFAR10-ResNet}
		\label{figsa1:e}
	\end{subfigure}   
	~ 
	\begin{subfigure}[b]{.45\textwidth}
		\includegraphics[width=\columnwidth]{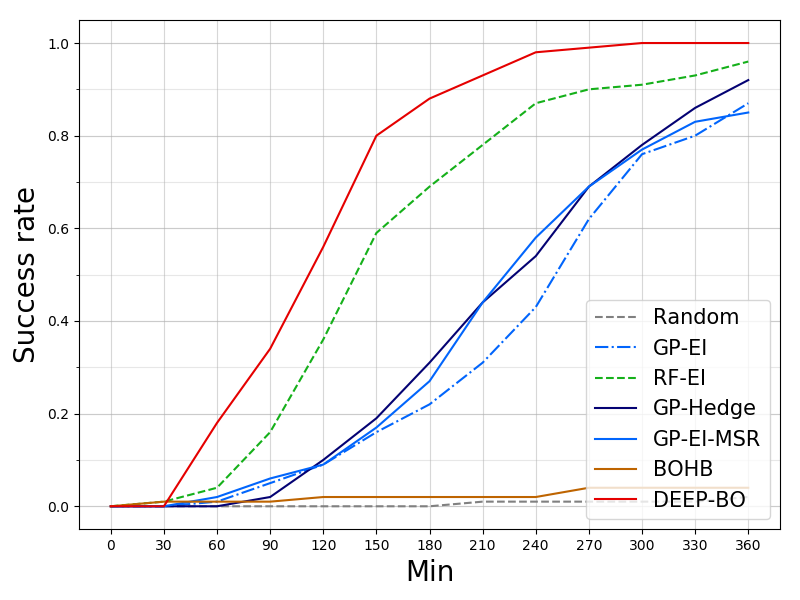}
		\caption{CIFAR100-VGG}
		\label{figs1:f}
	\end{subfigure}
	\caption*{Figure D1. \textit{Success rate} to achieve over the top-10 error performance using a single processor(higher is better). 
	}\label{figs1}
\end{figure}

\begin{figure}[H]
	\centering
	\begin{subfigure}[b]{.45\textwidth}
		\includegraphics[width=\columnwidth]{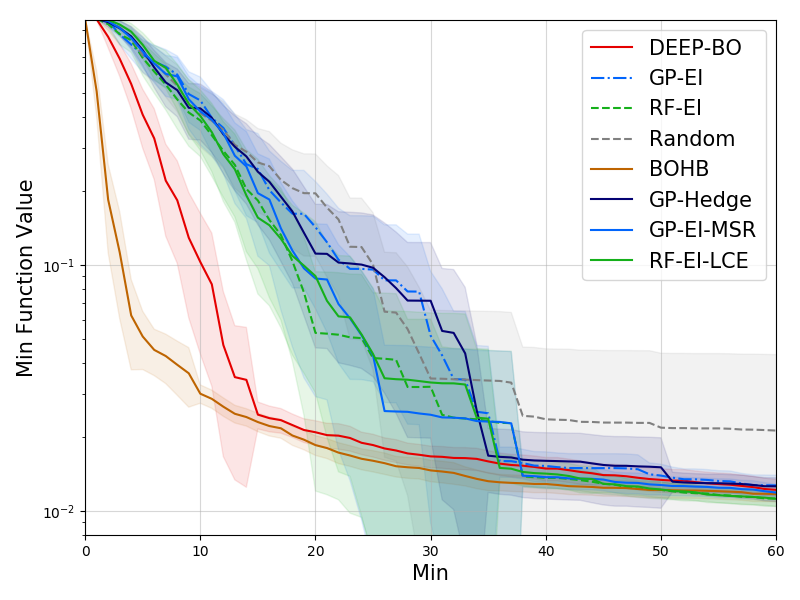}
		\caption{\char`\~1h}
		\label{figs2:1a}
	\end{subfigure}
	~ 
	\begin{subfigure}[b]{.45\textwidth}
		\includegraphics[width=\columnwidth]{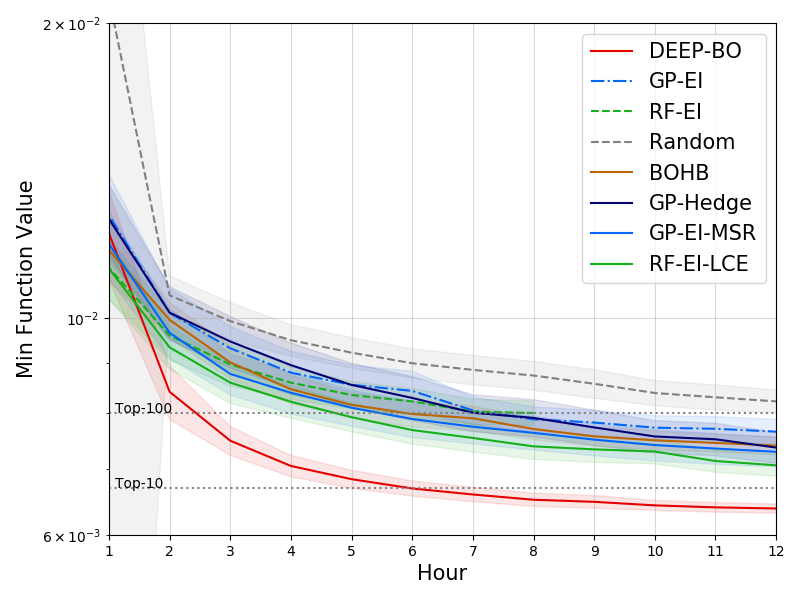}
		\caption{1\char`\~12h}
		\label{figs2:1b}
	\end{subfigure}
	\caption*{Figure D2-1. HPO performance comparison of MNIST-LeNet1 benchmark(lower is better). Note that the shaded areas are 0.25 $\sigma$ error bars.}
	\label{figs21}
\end{figure}

\begin{figure}[H]
	\centering
	\begin{subfigure}[b]{.45\textwidth}
		\includegraphics[width=\columnwidth]{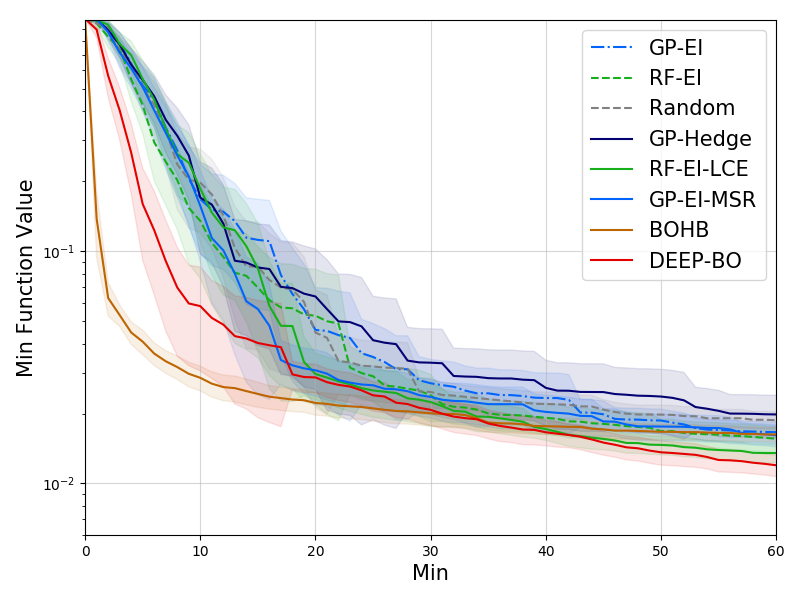}
		\caption{\char`\~1h}
		\label{figs2:2a}
	\end{subfigure}
	~ 
	\begin{subfigure}[b]{.45\textwidth}
		\includegraphics[width=\columnwidth]{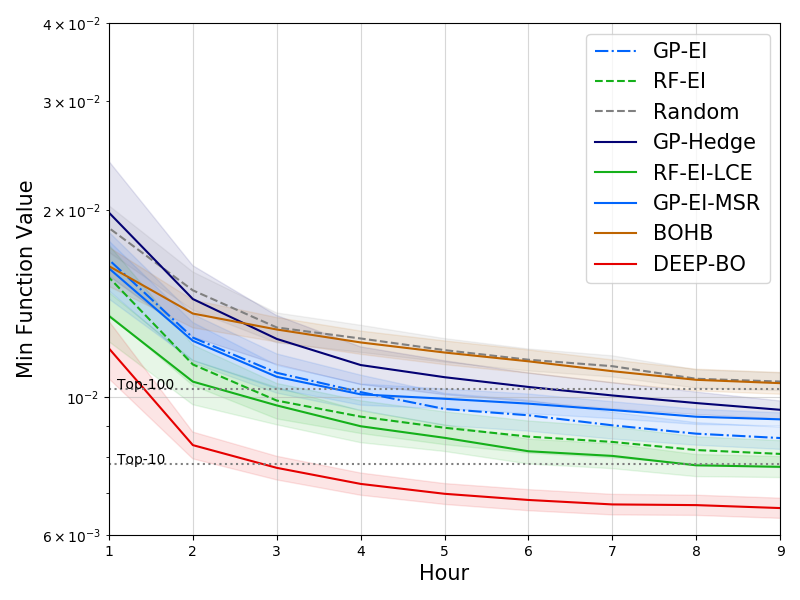}
		\caption{1\char`\~9h}
		\label{figs2:2b}
	\end{subfigure}
	
	\caption*{Figure D2-2. HPO performance of MNIST-LeNet2 benchmark. }
	\label{figs22}
\end{figure}

\begin{figure}[H]
	\centering
	\begin{subfigure}[b]{.45\textwidth}
		\includegraphics[width=\columnwidth]{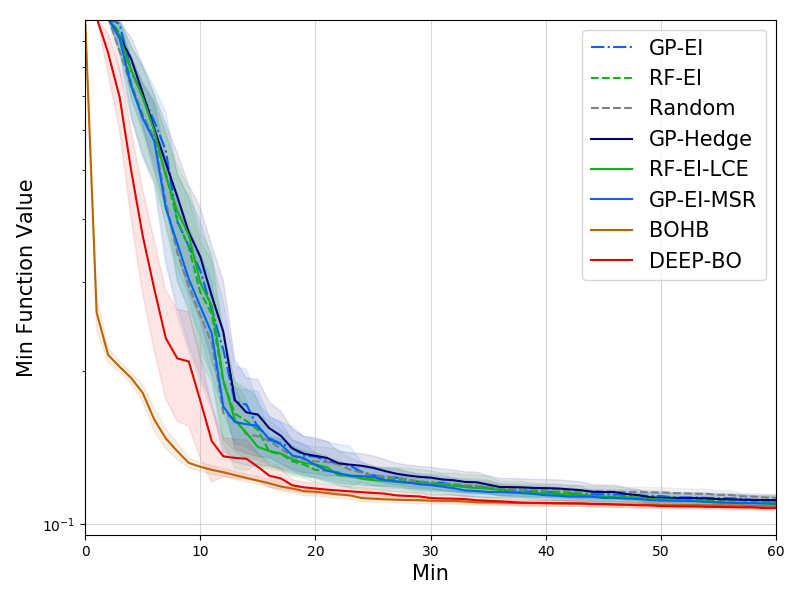}
		\caption{\char`\~1h}
		\label{figs2:3a}
	\end{subfigure}
	~ 
	\begin{subfigure}[b]{.45\textwidth}
		\includegraphics[width=\columnwidth]{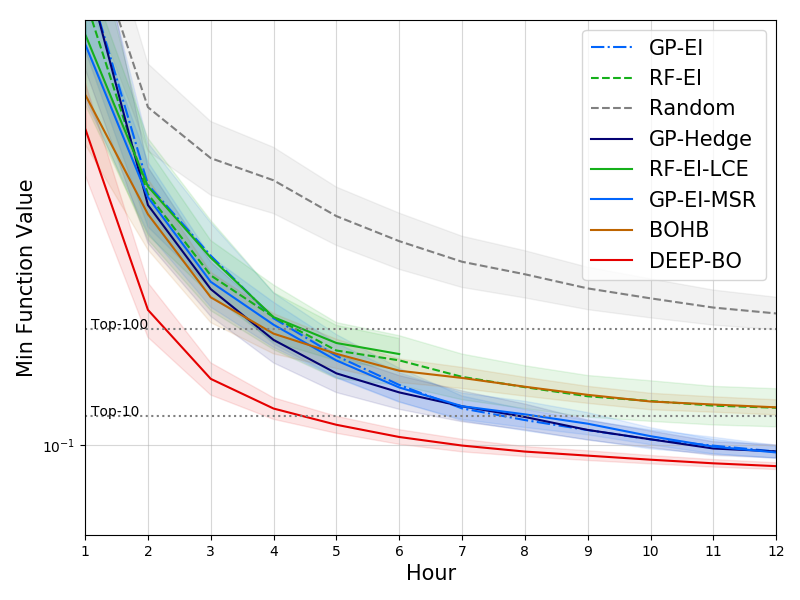}
		\caption{1\char`\~12h}
		\label{figs2:3b}
	\end{subfigure}
	
	\caption*{Figure D2-3. HPO performance of PTB-LSTM benchmark. }
	\label{figs23}
\end{figure}

\begin{figure}[H]
	\centering
	\begin{subfigure}[b]{.45\textwidth}
		\includegraphics[width=\columnwidth]{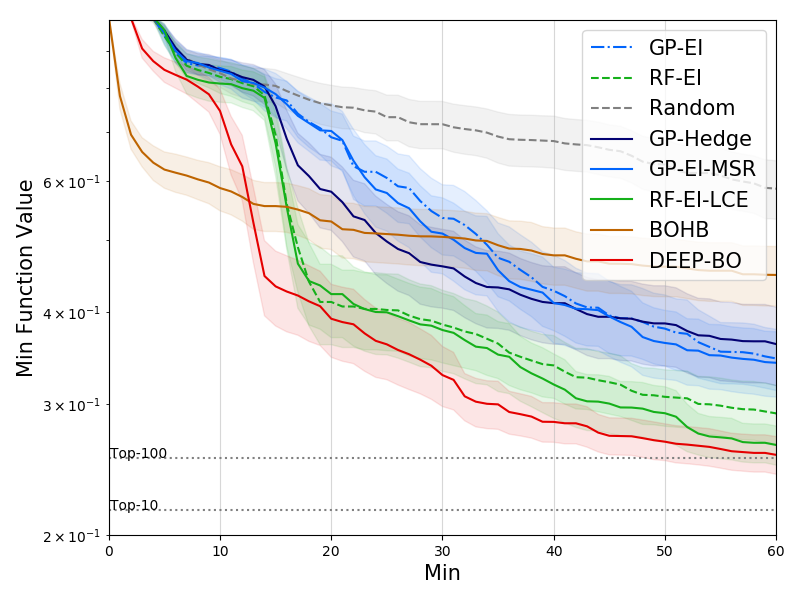}
		\caption{\char`\~1h}
		\label{figs2:4a}
	\end{subfigure}
	~ 
	\begin{subfigure}[b]{.45\textwidth}
		\includegraphics[width=\columnwidth]{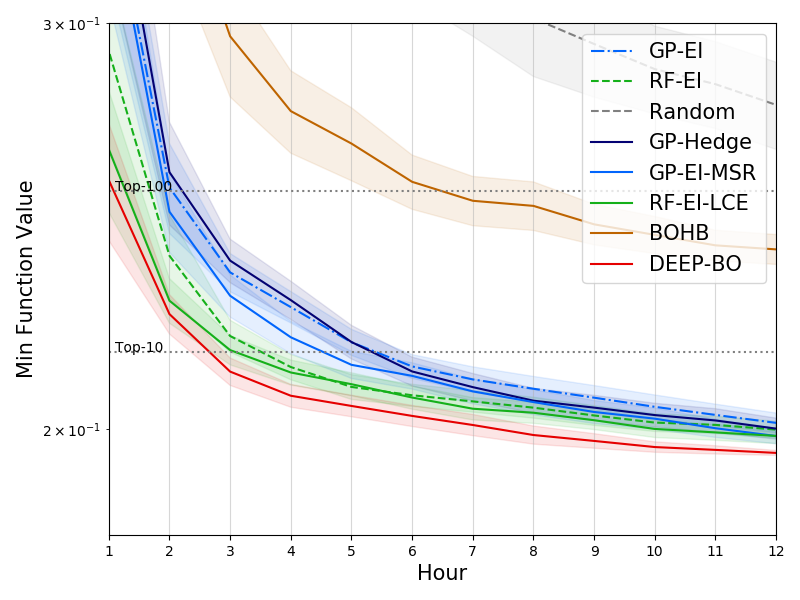}
		\caption{1\char`\~12h}
		\label{figs2:4b}
	\end{subfigure}
	
	\caption*{Figure D2-4. HPO performance of CIFAR10-VGG benchmark.}
	\label{figs24}
\end{figure}

\begin{figure}[H]
	\centering
	
	\begin{subfigure}[b]{.45\textwidth}
		\includegraphics[width=\columnwidth]{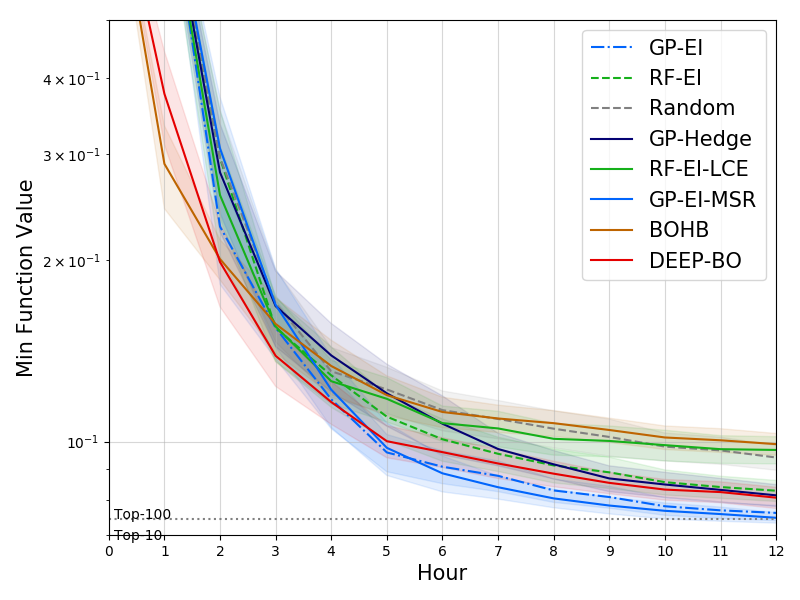}
		\caption{\char`\~12h}
		\label{figs2:5a}
	\end{subfigure}
	~ 
	\begin{subfigure}[b]{.45\textwidth}
		\includegraphics[width=\columnwidth]{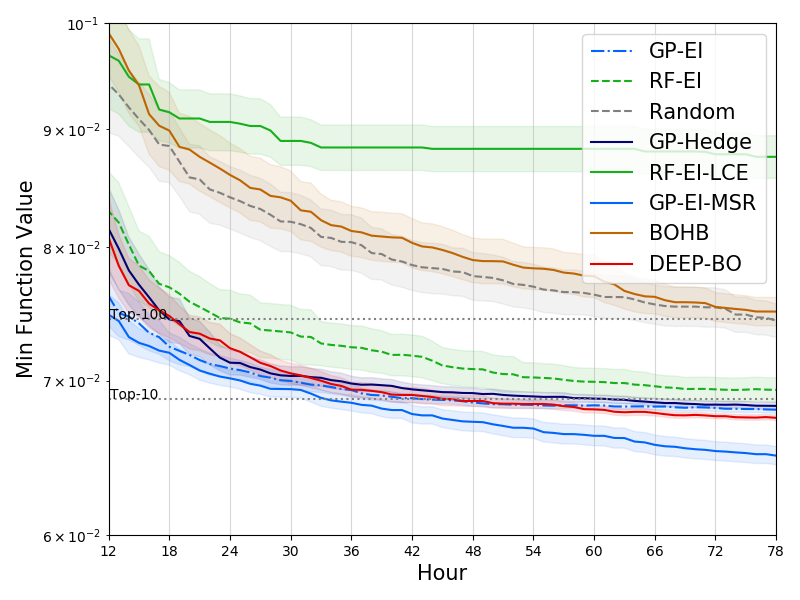}
		\caption{12\char`\~78h}
		\label{figs2:5b}
	\end{subfigure}
	
	\caption*{Figure D2-5. HPO performance of CIFAR10-ResNet benchmark.}
	\label{figs2}
\end{figure}

\begin{figure}[H]
	\centering
	\begin{subfigure}[b]{.45\textwidth}
		\includegraphics[width=\columnwidth]{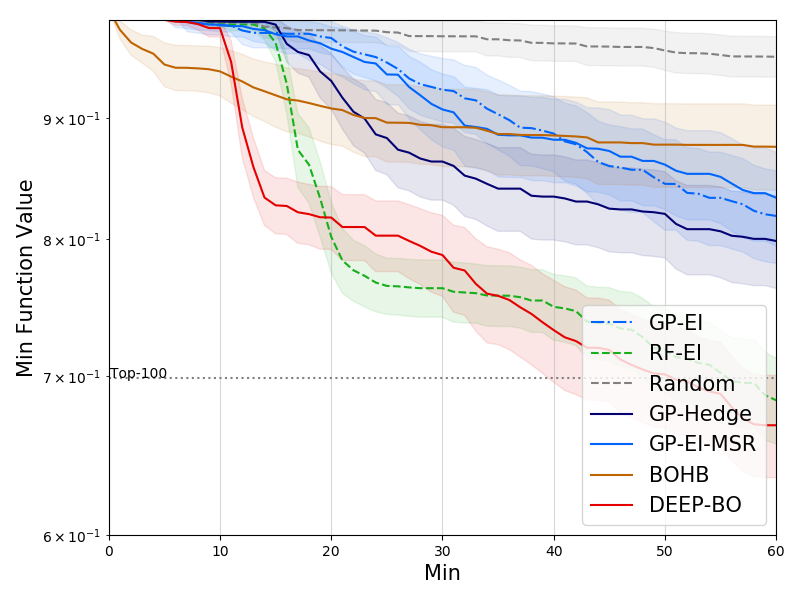}
		\caption{\char`\~1h}
		\label{figs2:6a}
	\end{subfigure} 
	~ 
	\begin{subfigure}[b]{.45\textwidth}
		\includegraphics[width=\columnwidth]{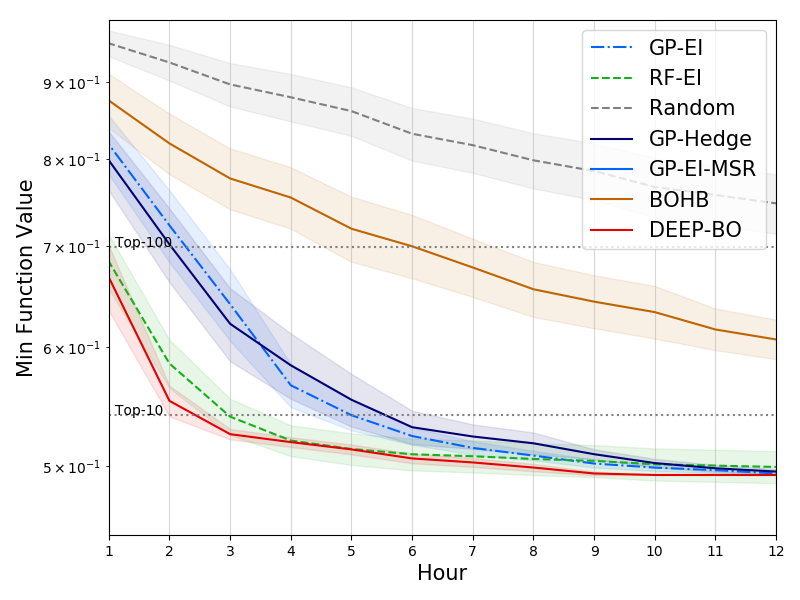}
		\caption{1\char`\~12h}
		\label{figs2:6b}
	\end{subfigure}
	
	\caption*{Figure D2-6. HPO performance of CIFAR100-VGG benchmark. }
	\label{figs25}
\end{figure}

\subsection*{Using 6 parallel processors}

\begin{table*}
	\centering
	\caption*{Table D3. Summary of the 6 processors performance achieving top-10 accuracy.} 
	\label{tab-b4}
	\resizebox{\columnwidth}{!}{
		\begin{tabular}{l|p{3.5cm}|r|r|r|r|r|r|r|r}
			\toprule
			\textbf{Measure} & \textbf{Benckmark} & \textbf{GP-EI} & \textbf{GP-PI} & \textbf{GP-UCB} & \textbf{RF-EI} & \textbf{RF-PI} & \textbf{RF-UCB} & \textbf{GP-EI-MSR} & \textbf{DEEP-BO} \\ 
			\midrule
			\multirow{7}{*}{\begin{tabular}[c]{@{}l@{}}Success\\     rate\end{tabular}}    
			& MNIST-LeNet1 \space\space\space\space\space(2h)    & 15\%           & 9\%           & 13\%            & 32\%           & 58\%           & 30\%            & 15\%  & \textbf{80\%}   \\  
			&  MNIST-LeNet2 \space\space\space\space\space(2h)     & 39\%           & 17\%           & 38\%             & 44\%           & 56\%           & 46\%            & 41\%  & \textbf{95\%}   \\  
			& PTB-LSTM \space\space\space\space\space\space\space\space\space\space\space(2h)     & 45\%           & 68\%           & 52\%            & 24\%           & 63\%           & 35\%            & 60\%  & \textbf{95\%}  \\  
			& CIFAR10-VGG \space\space\space\space\space(1h)         & 32\%           & 43\%           & 39\%            & 40\%           & 48\%           & 32\%            & 50\%  & \textbf{75\%}   \\  
			& CIFAR10-ResNet (15h)    & 61\%           & 59\%           & 70\%            & 37\%           & 58\%           & 55\%            & 81\%    & \textbf{99\%}     \\  
			& CIFAR100-VGG \space\space\space(1h)  & 30\%           & 25\%           & 37\%            & 51\%           & 59\%           & 47\%            & 43\%  & \textbf{99\%}  \\ 
			\cmidrule(l){2-10}
			& Mean  & 37\%           & 37\%           & 42\%            & 38\%           & 57\%           & 41\%            & 48\%    & \textbf{91\%}    \\ 
			
			\midrule
			\multirow{7}{*}{\begin{tabular}[c]{@{}l@{}}Expected\\     time \\(hour)\end{tabular}} 
			
			& MNIST-LeNet1       & $10.4\pm8.5$           & $11.9\pm8.9$           & $8.9\pm7.8$            & $4.7\pm5.2$           & $2.4\pm2.0$           & $4.6\pm4.3$            & $4.0\pm1.5$     & $1.4\pm1.0$    \\  
			&  MNIST-LeNet2       & $4.6\pm5.4$           & $8.4\pm7.3$           & $3.8\pm3.8$             & $3.0\pm2.8$           & $3.3\pm4.5$           & $2.4\pm1.9$            & $2.4\pm1.4$     & $0.8\pm1.0$    \\  
			& PTB-LSTM       & $2.1\pm0.7$           & $1.7\pm0.8$          & $2.0\pm0.7$            & $2.5\pm0.7$           & $1.8\pm0.9$           & $2.3\pm0.8$            & $1.8\pm0.8$    & $1.0\pm0.4$  \\  
			& CIFAR10-VGG         & $1.5\pm0.9$           & $1.5\pm1.1$           & $1.5\pm1.0$            & $1.5\pm1.0$           & $1.4\pm1.0$           & $1.8\pm1.1$            & $1.3\pm0.8$     & $0.8\pm0.4$  \\  
			& CIFAR10-ResNet    & $13.0\pm5.5$           & $14.1\pm6.6$           & $12.3\pm5.7$            & $18.1\pm7.3$           & $14.4\pm6.6$           & $14.6\pm6.6$            & $11.4\pm4.1$      & $7.7\pm2.5$    \\  
			& CIFAR100-VGG    & $1.2\pm0.5$           & $1.4\pm0.5$           & $1.2\pm0.5$            & $1.1\pm0.4$           & $1.1\pm1.0$           & $1.0\pm0.4$            & $1.1\pm0.4$ & $0.5\pm0.2$  \\
			\cmidrule(l){2-10}
			& Mean    & 5.5           & 6.5           & 5.0            & 5.2           & 4.1           & 4.4            & 3.7   & \textbf{2.0} 	\\
			\bottomrule
		\end{tabular}		
	}
\end{table*}

As shown in Table D3, our algorithm can complete the tasks in merely 2 hours on average using 6 processors. 
On the mean performance of all benchmarks, the \textit{success rate} of DEEP-BO was 34\% superior compared to RF-PI and the \textit{expected time} decreased by 2.1 hours. 
Regarding \textit{success rate}, DEEP-BO achieved 117\%, 88\% of average improvements over the individual BO algorithms and ETR-applied algorithm, respectively. 
Regarding \textit{expected time}, DEEP-BO achieved 60\%, 44\% of time reductions over the individual algorithms or the ETR-applied algorithm, respectively.

\begin{figure}[H]
	\centering
	\begin{subfigure}[b]{.45\textwidth}
		\includegraphics[width=\columnwidth]{figs/fig-mnist2-p1.png}
		\caption{MNIST-LeNet1}
		\label{figs3:a}
	\end{subfigure}
	~ 
	\begin{subfigure}[b]{.45\textwidth}
		\includegraphics[width=\columnwidth]{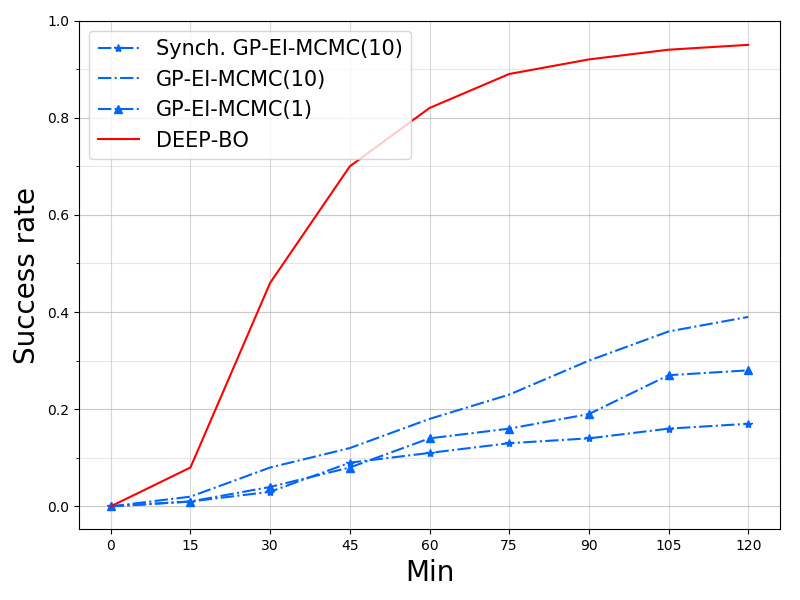}
		\caption{MNIST-LeNet2}
		\label{figs3:b}
	\end{subfigure}
	~ 
	\begin{subfigure}[b]{.45\textwidth}
		\includegraphics[width=\columnwidth]{figs/fig-ptb-p1.png}
		\caption{PTB-LSTM}
		\label{figs3:c}
	\end{subfigure}
	~ 
	\begin{subfigure}[b]{.45\textwidth}
		\includegraphics[width=\columnwidth]{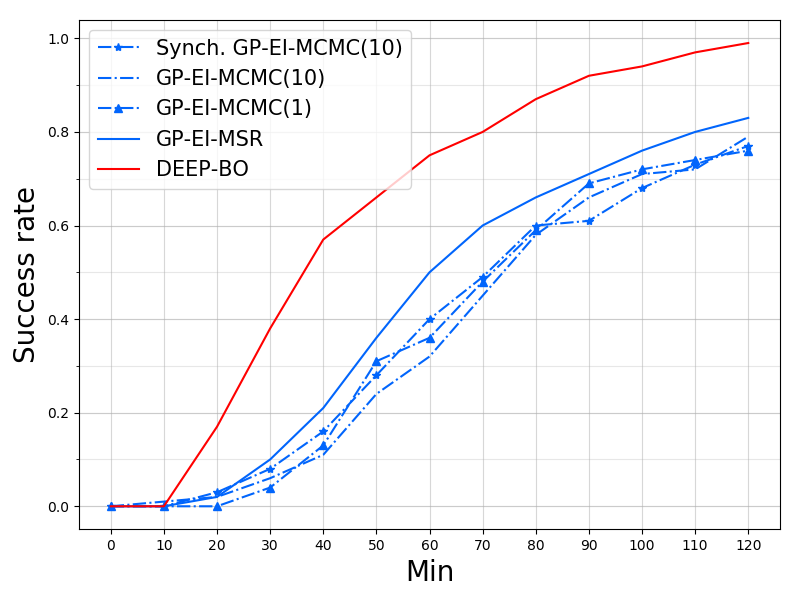}
		\caption{CIFAR10-VGG}
		\label{figs3:d}
	\end{subfigure}
	~ 
	\begin{subfigure}[b]{.45\textwidth}
		\includegraphics[width=\columnwidth]{figs/fig-cifar10-resnet-p1.png}
		\caption{CIFAR10-ResNet}
		\label{figs3:e}
	\end{subfigure}   
	~ 
	\begin{subfigure}[b]{.45\textwidth}
		\includegraphics[width=\columnwidth]{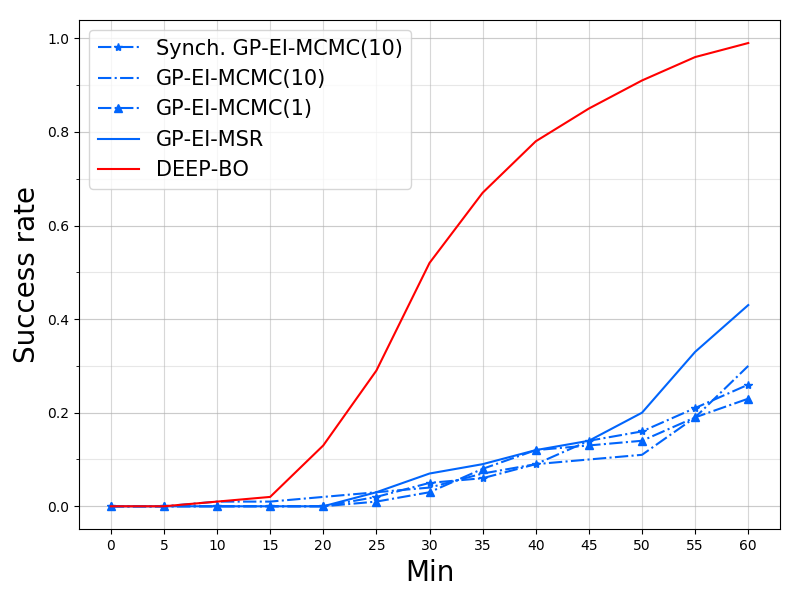}
		\caption{CIFAR100-VGG}
		\label{figs3:f}
	\end{subfigure}      
	\caption*{Figure D3. \textit{Success rate} to achieve over the top-10 error performance when using six processors.}
	\label{figs3}
\end{figure}

\subsection*{Extra comparisons}


For fair comparison with HB, we have run the regression sample provided by a HB implementation. 
See \url{https://github.com/zygmuntz/hyperband/blob/master/defs_regression/keras_mlp.py}

\begin{figure*}[!ht]
	\centering
	\begin{subfigure}[b]{.9\textwidth}
		\includegraphics[width=\columnwidth]{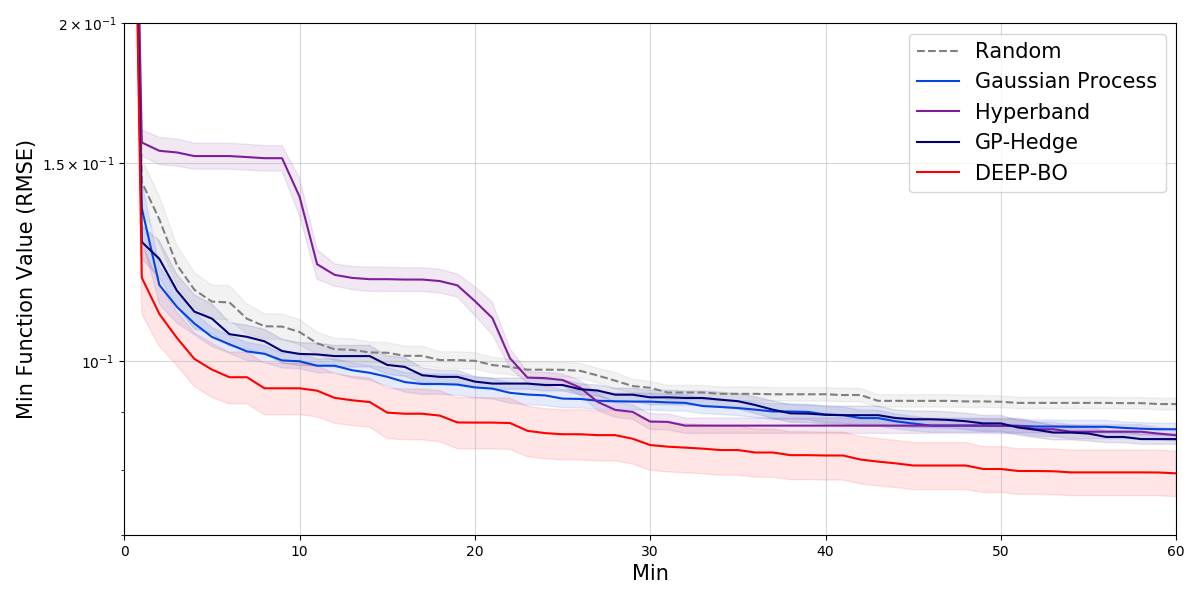}
		\label{figc3:a}
	\end{subfigure}
	
	\caption*{Figure D4. HPO of Regression model on kin8nm dataset.}
	\label{figc3}
\end{figure*} 

For fair comparison with BOHB, we have run the sample code provided by BOHB's author.
This example aims to demonstrate how to use the BOHB to tune the hyperparameter of the classification model learned with MNIST data.
Its hyperparameter space has 9-dimensional space which consists of 5 discrete, 3 continuous, and 1 categorical variable respectively. See \url{https://github.com/automl/HpBandSter/blob/master/hpbandster/examples/example_5_keras_worker.py} for details.

\begin{figure*}[!ht]
	\centering
	\begin{subfigure}[b]{.3\textwidth}
		\includegraphics[width=\columnwidth]{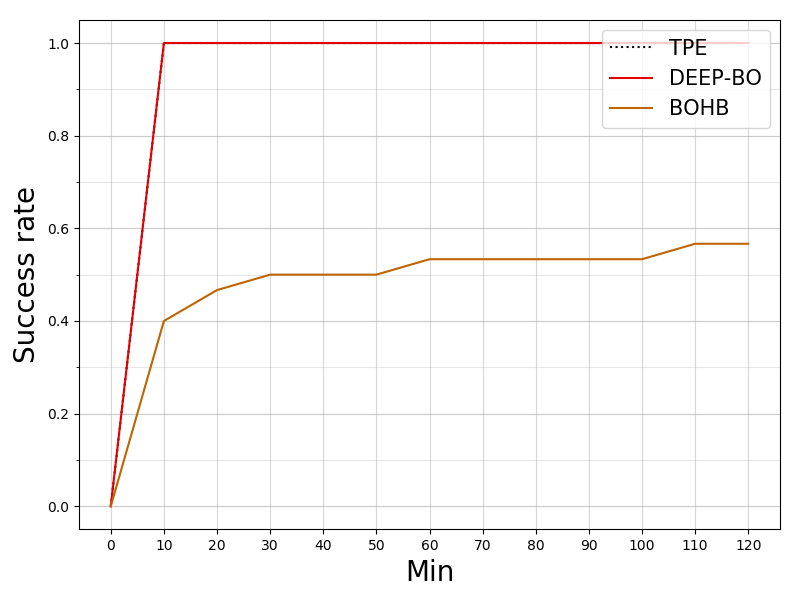}
		\caption{Easy target (5\% error)}
		\label{figc3:a}
	\end{subfigure}
	\begin{subfigure}[b]{.3\textwidth}
		\includegraphics[width=\columnwidth]{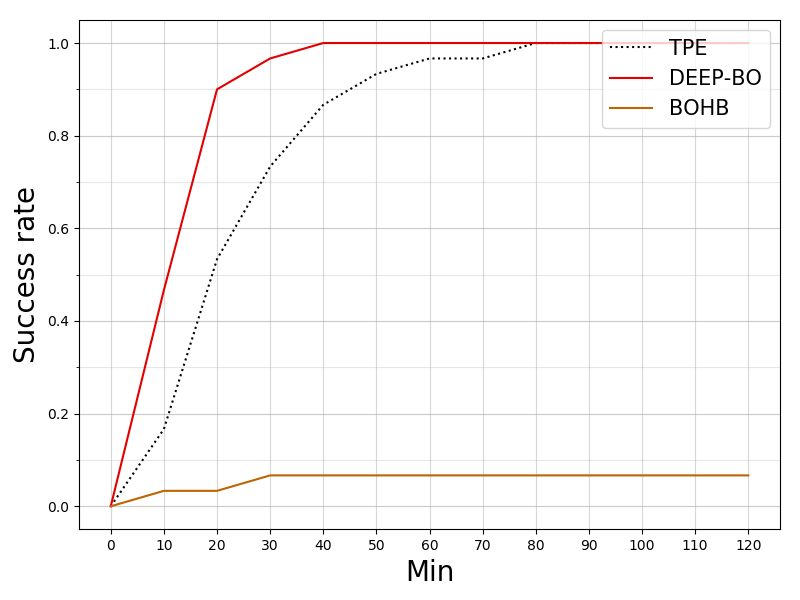}
		\caption{Hard target (1\% error)}
		\label{figc3:b}
	\end{subfigure}
	\begin{subfigure}[b]{.3\textwidth}
		\includegraphics[width=\columnwidth]{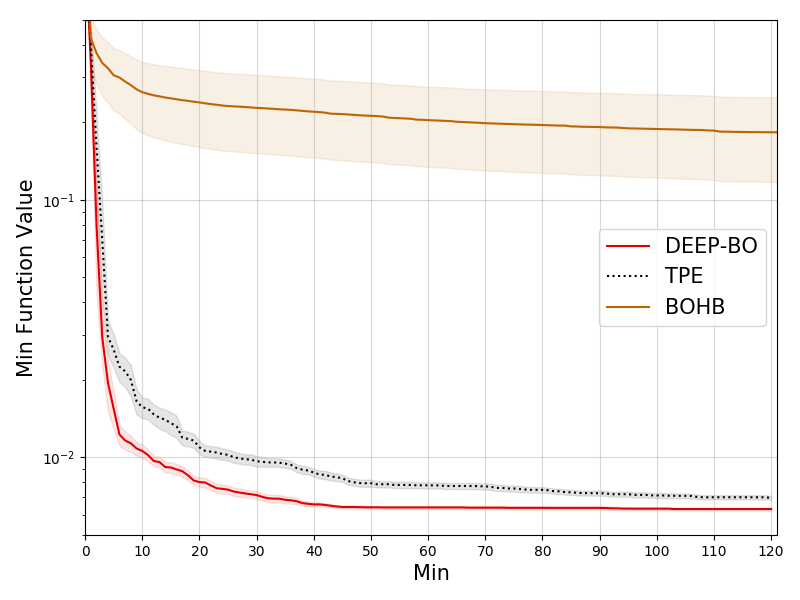}
		\caption{Best error performance}
		\label{figc3:c}
	\end{subfigure}
	\caption*{Figure D5. Performance comparison on a benchmark problem provided by BOHB.}
	\label{figc3}
\end{figure*} 

As shown in Figure D5, our DEEP-BO outperforms BOHB and TPE. 
However, BOHB could not achieve any better result compared with TPE. 
This experiment was repeated 30 times to get a statistically significant result(F-statistics is 13.18, p-value is 1e-05).
We will discuss later why BOHB failed catastrophically in DNN problem.

\subsection*{Performance on the number of hyperparameters}

\cite{eggensperger2013towards} summarize the performance characteristics of 3 HPO algorithms(Spearmint, SMAC, TPE) which are used widely. Spearmint which models using GP tends to be better in the low-dimensional continuous problem. Instead, SMAC and TPE tend to be better in the high-dimensional problem. 
Afterall, Figure C6 shows that our diversification strategy always performs well at any dimensional problem. 

\begin{figure*}[!ht]
	\centering
	\begin{subfigure}[b]{.45\textwidth}
		\includegraphics[width=\columnwidth]{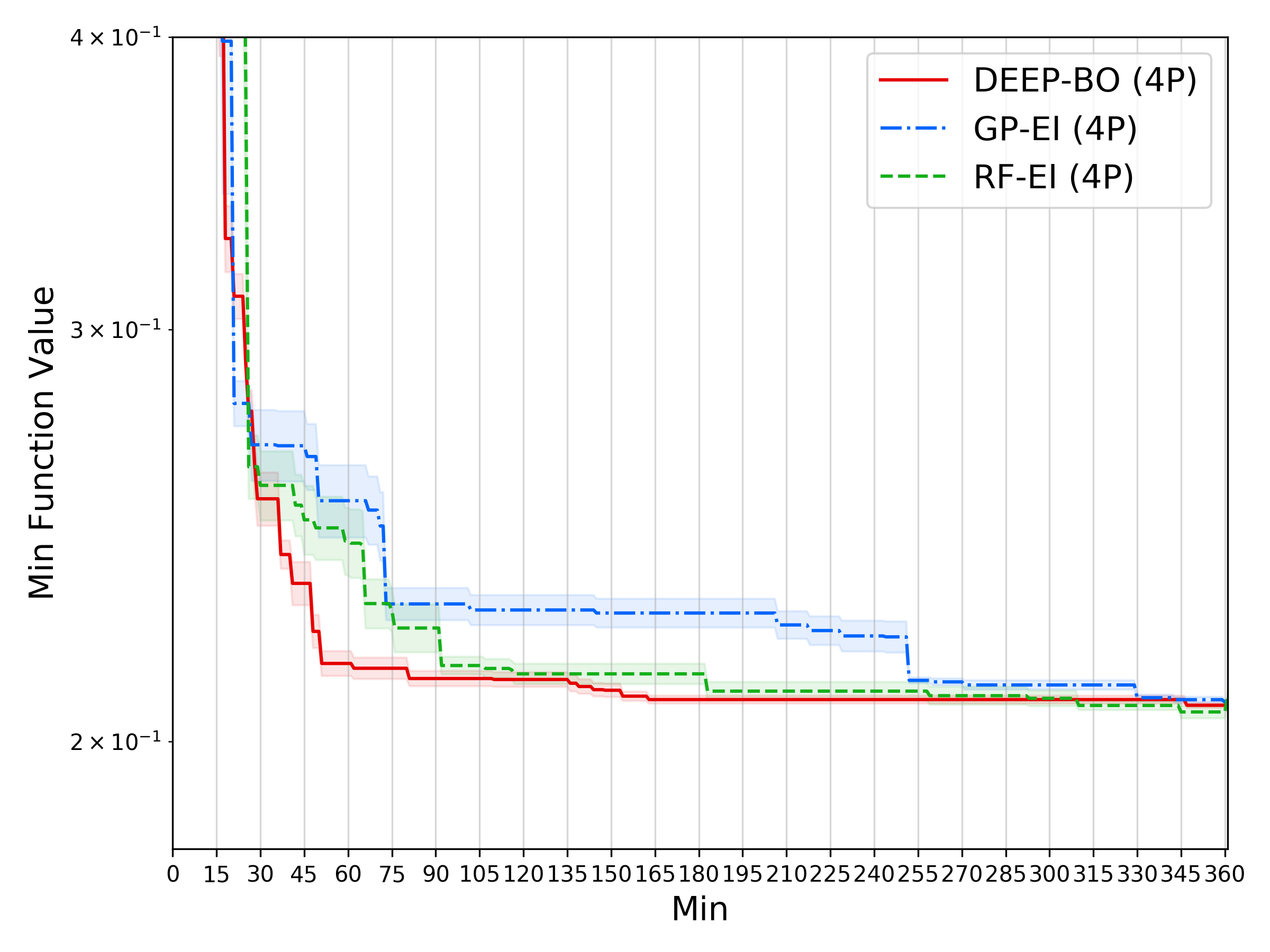}
		\caption{4 continuous parameters}
		\label{figd1:a}
	\end{subfigure}
	\begin{subfigure}[b]{.45\textwidth}
		\includegraphics[width=\columnwidth]{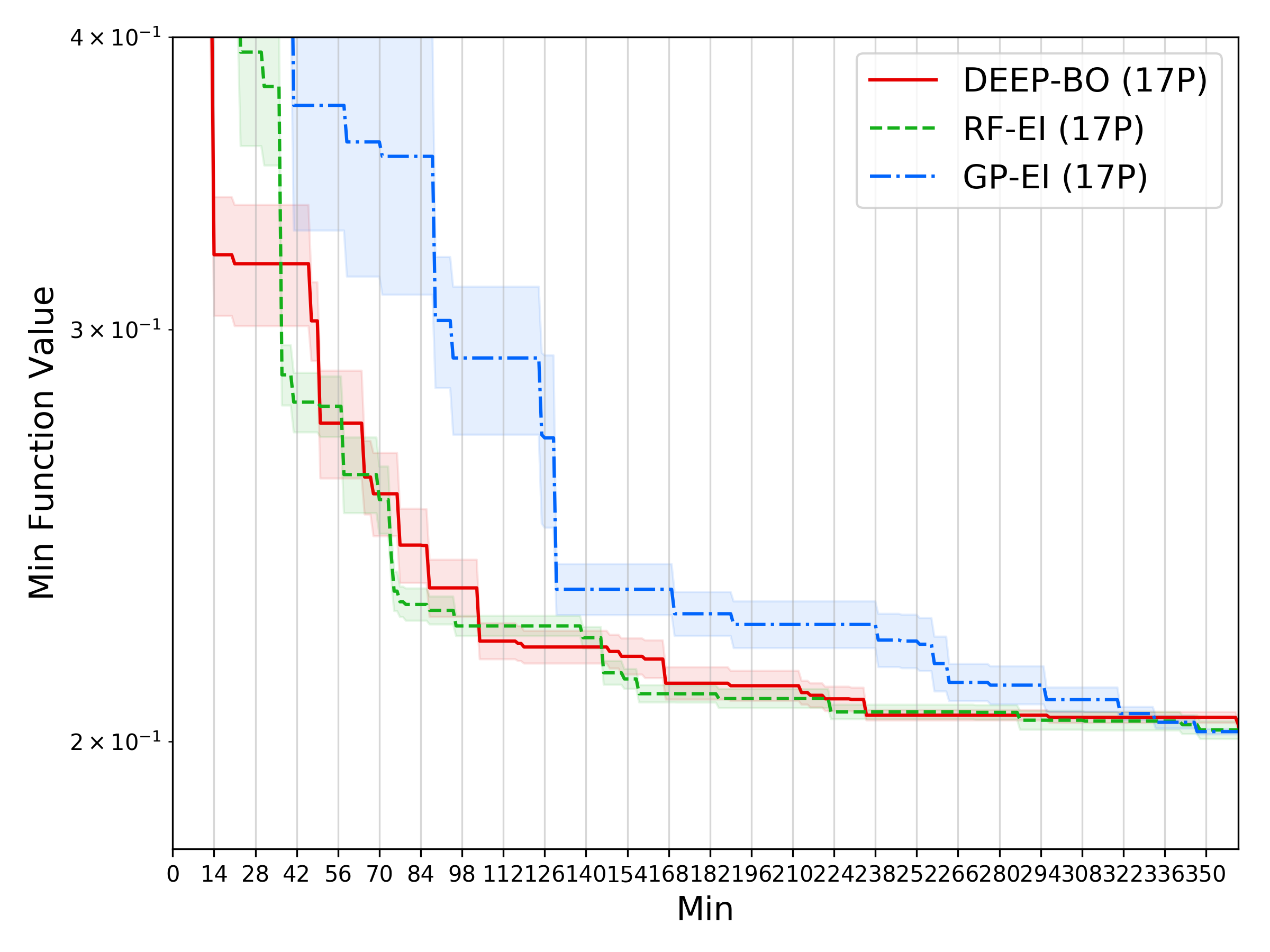}
		\caption{17 discrete, continuous, categorical parameters}
		\label{figd1:b}
	\end{subfigure}
	\caption*{Figure D6. Performance comparisons when optimizing the different dimensional problem. We created this problem to evaluate BO algorithms only for the dimension size issue by decreasing or increasing the number of hyperparameters of CIFAR10-VGG benchmark problem.}
	\label{figd1}
\end{figure*}

\subsection*{Performance by max epochs setting}

\begin{figure}[!ht]
	\centering
	\begin{subfigure}[b]{.95\textwidth}
		\includegraphics[width=\columnwidth]{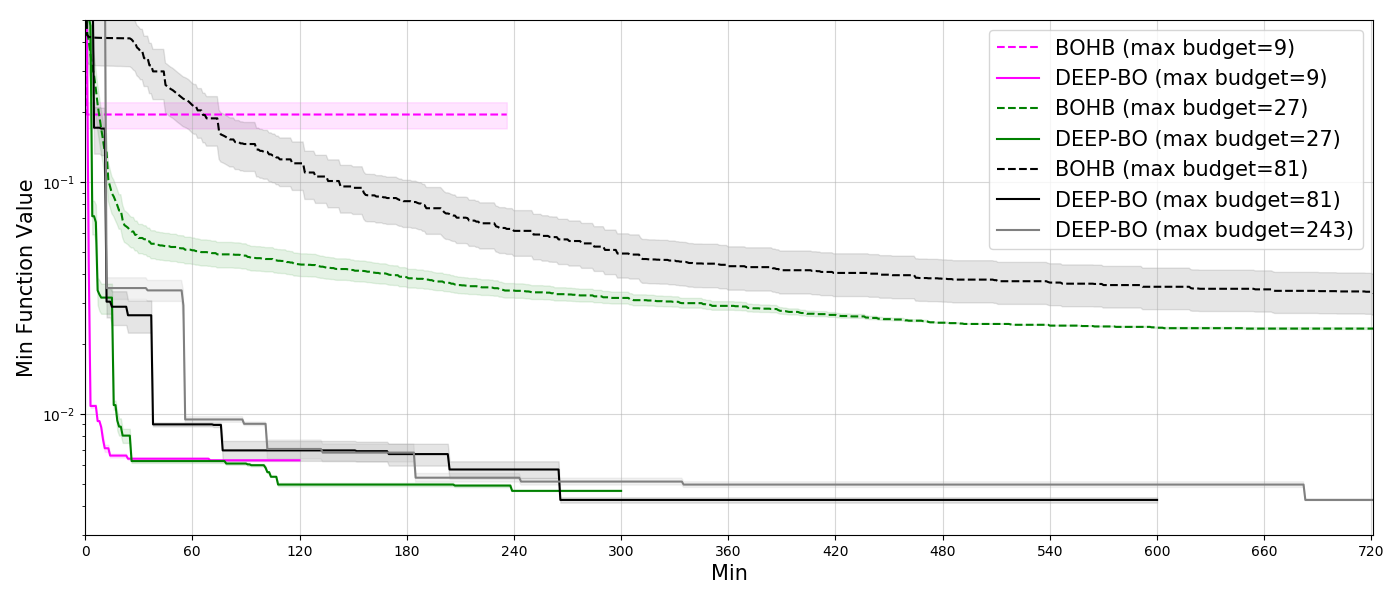}
		\caption*{
			Figure D7. Performance comparisons when the number of max epochs is increased. We evaluated our algorithm on the sample code provided by BOHB's author. 
Our DEEP-BO always perform better than BOHB.
In the case of this problem, the asymptotic performance tends to achieved when 6 hours later.
Too small or too much max epoch tends to be ineffective to achieve the target goal in a given budget.
	}
		\label{figsc3}
	\end{subfigure}
\end{figure}

In DNN HPO problems, setting proper max epochs of training is very important to find their optimal performance. 
Because of no way of predicting the optimal number of epochs before evaluated, this remains an art of human experts.
If we have a lack of prior knowledge about the problem to optimize, too few or too much of max epochs tends to bring unexpected results even after spending much expense. Therefore, escalating the target performance is required by considering the budget.

%% file: main_arxiv.bbl
\begin{thebibliography}{34}
\providecommand{\natexlab}[1]{#1}
\providecommand{\url}[1]{\texttt{#1}}
\expandafter\ifx\csname urlstyle\endcsname\relax
  \providecommand{\doi}[1]{doi: #1}\else
  \providecommand{\doi}{doi: \begingroup \urlstyle{rm}\Url}\fi

\bibitem[Baker et~al.(2017)Baker, Gupta, Raskar, and
  Naik]{baker2017accelerating}
B.~Baker, O.~Gupta, R.~Raskar, and N.~Naik.
\newblock Accelerating neural architecture search using performance prediction.
\newblock \emph{arXiv preprint arXiv:1705.10823}, 2017.

\bibitem[Bergstra and Bengio(2012)]{bergstra2012random}
J.~Bergstra and Y.~Bengio.
\newblock Random search for hyper-parameter optimization.
\newblock \emph{Journal of Machine Learning Research}, 13\penalty0
  (Feb):\penalty0 281--305, 2012.

\bibitem[Bergstra et~al.(2011)Bergstra, Bardenet, Bengio, and
  K{\'e}gl]{bergstra2011algorithms}
J.~S. Bergstra, R.~Bardenet, Y.~Bengio, and B.~K{\'e}gl.
\newblock Algorithms for hyper-parameter optimization.
\newblock In \emph{Advances in neural information processing systems}, pages
  2546--2554, 2011.

\bibitem[Choi et~al.(2018)Choi, Cho, and Rhee]{choi2018difficulty}
D.~Choi, H.~Cho, and W.~Rhee.
\newblock On the difficulty of dnn hyperparameter optimization using learning
  curve prediction.
\newblock In \emph{TENCON 2018-2018 IEEE Region 10 Conference}, pages
  0651--0656. IEEE, 2018.

\bibitem[Contal et~al.(2013)Contal, Buffoni, Robicquet, and
  Vayatis]{contal2013parallel}
E.~Contal, D.~Buffoni, A.~Robicquet, and N.~Vayatis.
\newblock Parallel gaussian process optimization with upper confidence bound
  and pure exploration.
\newblock In \emph{Joint European Conference on Machine Learning and Knowledge
  Discovery in Databases}, pages 225--240. Springer, 2013.

\bibitem[Desautels et~al.(2014)Desautels, Krause, and
  Burdick]{desautels2014parallelizing}
T.~Desautels, A.~Krause, and J.~W. Burdick.
\newblock Parallelizing exploration-exploitation tradeoffs in gaussian process
  bandit optimization.
\newblock \emph{The Journal of Machine Learning Research}, 15\penalty0
  (1):\penalty0 3873--3923, 2014.

\bibitem[Domhan et~al.(2015)Domhan, Springenberg, and
  Hutter]{domhan2015speeding}
T.~Domhan, J.~T. Springenberg, and F.~Hutter.
\newblock Speeding up automatic hyperparameter optimization of deep neural
  networks by extrapolation of learning curves.
\newblock In \emph{Proceedings of the 24th International Joint Conference on
  Artificial Intelligence (IJCAI)}, 2015.

\bibitem[Eggensperger et~al.(2013)Eggensperger, Feurer, Hutter, Bergstra,
  Snoek, Hoos, and Leyton-Brown]{eggensperger2013towards}
K.~Eggensperger, M.~Feurer, F.~Hutter, J.~Bergstra, J.~Snoek, H.~Hoos, and
  K.~Leyton-Brown.
\newblock Towards an empirical foundation for assessing bayesian optimization
  of hyperparameters.
\newblock In \emph{NIPS workshop on Bayesian Optimization in Theory and
  Practice}, pages 1--5, 2013.

\bibitem[Eggensperger et~al.(2015)Eggensperger, Hutter, Hoos, and
  Leyton-Brown]{eggensperger2015efficient}
K.~Eggensperger, F.~Hutter, H.~H. Hoos, and K.~Leyton-Brown.
\newblock Efficient benchmarking of hyperparameter optimizers via surrogates.
\newblock In \emph{AAAI}, pages 1114--1120, 2015.

\bibitem[Falkner et~al.(2018)Falkner, Klein, and Hutter]{falkner2018bohb}
S.~Falkner, A.~Klein, and F.~Hutter.
\newblock Bohb: Robust and efficient hyperparameter optimization at scale.
\newblock \emph{arXiv preprint arXiv:1807.01774}, 2018.

\bibitem[Golovin et~al.(2017)Golovin, Solnik, Moitra, Kochanski, Karro, and
  Sculley]{golovin2017google}
D.~Golovin, B.~Solnik, S.~Moitra, G.~Kochanski, J.~Karro, and D.~Sculley.
\newblock Google vizier: A service for black-box optimization.
\newblock In \emph{Proceedings of the 23rd ACM SIGKDD International Conference
  on Knowledge Discovery and Data Mining}, pages 1487--1495. ACM, 2017.

\bibitem[Gonz{\'a}lez et~al.(2016)Gonz{\'a}lez, Dai, Hennig, and
  Lawrence]{gonzalez2016batch}
J.~Gonz{\'a}lez, Z.~Dai, P.~Hennig, and N.~Lawrence.
\newblock Batch bayesian optimization via local penalization.
\newblock In \emph{Artificial Intelligence and Statistics}, pages 648--657,
  2016.

\bibitem[G{\"u}l{\c{c}}ehre and Bengio(2016)]{gulccehre2016knowledge}
{\c{C}}.~G{\"u}l{\c{c}}ehre and Y.~Bengio.
\newblock Knowledge matters: Importance of prior information for optimization.
\newblock \emph{The Journal of Machine Learning Research}, 17\penalty0
  (1):\penalty0 226--257, 2016.

\bibitem[Hoffman et~al.(2011)Hoffman, Brochu, and
  de~Freitas]{hoffman2011portfolio}
M.~D. Hoffman, E.~Brochu, and N.~de~Freitas.
\newblock Portfolio allocation for bayesian optimization.
\newblock In \emph{UAI}, pages 327--336, 2011.

\bibitem[Hutter et~al.(2011)Hutter, Hoos, and
  Leyton-Brown]{hutter2011sequential}
F.~Hutter, H.~H. Hoos, and K.~Leyton-Brown.
\newblock Sequential model-based optimization for general algorithm
  configuration.
\newblock In \emph{International Conference on Learning and Intelligent
  Optimization}, pages 507--523. Springer, 2011.

\bibitem[Istrate et~al.(2018)Istrate, Scheidegger, Mariani, Nikolopoulos,
  Bekas, and Malossi]{istrate2018tapas}
R.~Istrate, F.~Scheidegger, G.~Mariani, D.~Nikolopoulos, C.~Bekas, and A.~C.~I.
  Malossi.
\newblock Tapas: Train-less accuracy predictor for architecture search.
\newblock \emph{arXiv preprint arXiv:1806.00250}, 2018.

\bibitem[Jones et~al.(1998)Jones, Schonlau, and Welch]{jones1998efficient}
D.~R. Jones, M.~Schonlau, and W.~J. Welch.
\newblock Efficient global optimization of expensive black-box functions.
\newblock \emph{Journal of Global optimization}, 13\penalty0 (4):\penalty0
  455--492, 1998.

\bibitem[Kathuria et~al.(2016)Kathuria, Deshpande, and
  Kohli]{kathuria2016batched}
T.~Kathuria, A.~Deshpande, and P.~Kohli.
\newblock Batched gaussian process bandit optimization via determinantal point
  processes.
\newblock In \emph{Advances in Neural Information Processing Systems}, pages
  4206--4214, 2016.

\bibitem[Klein et~al.(2016)Klein, Falkner, Springenberg, and
  Hutter]{klein2016learning}
A.~Klein, S.~Falkner, J.~T. Springenberg, and F.~Hutter.
\newblock Learning curve prediction with bayesian neural networks.
\newblock 2016.

\bibitem[Kushner(1964)]{kushner1964new}
H.~J. Kushner.
\newblock A new method of locating the maximum point of an arbitrary multipeak
  curve in the presence of noise.
\newblock \emph{Journal of Basic Engineering}, 86\penalty0 (1):\penalty0
  97--106, 1964.

\bibitem[Li et~al.(2017)Li, Jamieson, DeSalvo, Rostamizadeh, and
  Talwalkar]{li2017hyperband}
L.~Li, K.~Jamieson, G.~DeSalvo, A.~Rostamizadeh, and A.~Talwalkar.
\newblock Hyperband: A novel bandit-based approach to hyperparameter
  optimization.
\newblock \emph{The Journal of Machine Learning Research}, 18\penalty0
  (1):\penalty0 6765--6816, 2017.

\bibitem[Markowitz(1952)]{markowitz1952portfolio}
H.~Markowitz.
\newblock Portfolio selection.
\newblock \emph{The journal of finance}, 7\penalty0 (1):\penalty0 77--91, 1952.

\bibitem[Mockus et~al.(1978)Mockus, Tiesis, and Zilinskas]{mockus1978toward}
J.~Mockus, V.~Tiesis, and A.~Zilinskas.
\newblock chapter bayesian methods for seeking the extremum.
\newblock \emph{Toward global optimization, volume 2}, 1978.

\bibitem[Nguyen et~al.(2016)Nguyen, Rana, Gupta, Li, and
  Venkatesh]{nguyen2016budgeted}
V.~Nguyen, S.~Rana, S.~K. Gupta, C.~Li, and S.~Venkatesh.
\newblock Budgeted batch bayesian optimization.
\newblock In \emph{Data Mining (ICDM), 2016 IEEE 16th International Conference
  on}, pages 1107--1112. IEEE, 2016.

\bibitem[Prechelt(1998)]{prechelt1998automatic}
L.~Prechelt.
\newblock Automatic early stopping using cross validation: quantifying the
  criteria.
\newblock \emph{Neural Networks}, 11\penalty0 (4):\penalty0 761--767, 1998.

\bibitem[Shah and Ghahramani(2015)]{shah2015parallel}
A.~Shah and Z.~Ghahramani.
\newblock Parallel predictive entropy search for batch global optimization of
  expensive objective functions.
\newblock In \emph{Advances in Neural Information Processing Systems}, pages
  3330--3338, 2015.

\bibitem[Snoek et~al.(2012)Snoek, Larochelle, and Adams]{snoek2012practical}
J.~Snoek, H.~Larochelle, and R.~P. Adams.
\newblock Practical bayesian optimization of machine learning algorithms.
\newblock In \emph{Advances in neural information processing systems}, pages
  2951--2959, 2012.

\bibitem[Sobol'(1967)]{sobol1967distribution}
I.~M. Sobol'.
\newblock On the distribution of points in a cube and the approximate
  evaluation of integrals.
\newblock \emph{Zhurnal Vychislitel'noi Matematiki i Matematicheskoi Fiziki},
  7\penalty0 (4):\penalty0 784--802, 1967.

\bibitem[Springenberg et~al.(2016)Springenberg, Klein, Falkner, and
  Hutter]{springenberg2016bayesian}
J.~T. Springenberg, A.~Klein, S.~Falkner, and F.~Hutter.
\newblock Bayesian optimization with robust bayesian neural networks.
\newblock In \emph{Advances in Neural Information Processing Systems}, pages
  4134--4142, 2016.

\bibitem[Srinivas et~al.(2009)Srinivas, Krause, Kakade, and
  Seeger]{srinivas2009gaussian}
N.~Srinivas, A.~Krause, S.~M. Kakade, and M.~Seeger.
\newblock Gaussian process optimization in the bandit setting: No regret and
  experimental design.
\newblock \emph{arXiv preprint arXiv:0912.3995}, 2009.

\bibitem[Wang et~al.(2017)Wang, Gehring, Kohli, and Jegelka]{wang2017batched}
Z.~Wang, C.~Gehring, P.~Kohli, and S.~Jegelka.
\newblock Batched large-scale bayesian optimization in high-dimensional spaces.
\newblock \emph{arXiv preprint arXiv:1706.01445}, 2017.

\bibitem[Wolpert et~al.(1997)Wolpert, Macready, et~al.]{wolpert1997no}
D.~H. Wolpert, W.~G. Macready, et~al.
\newblock No free lunch theorems for optimization.
\newblock \emph{IEEE transactions on evolutionary computation}, 1\penalty0
  (1):\penalty0 67--82, 1997.

\bibitem[Wu and Frazier(2016)]{wu2016parallel}
J.~Wu and P.~Frazier.
\newblock The parallel knowledge gradient method for batch bayesian
  optimization.
\newblock In \emph{Advances in Neural Information Processing Systems}, pages
  3126--3134, 2016.

\bibitem[Zheng and Tse(2003)]{zheng2003diversity}
L.~Zheng and D.~N.~C. Tse.
\newblock Diversity and multiplexing: A fundamental tradeoff in
  multiple-antenna channels.
\newblock \emph{IEEE Transactions on information theory}, 49\penalty0
  (5):\penalty0 1073--1096, 2003.

\end{thebibliography}
